\documentclass[10pt,twocolumn,letterpaper]{article}
\pdfoutput=1
\usepackage{wacv}
\usepackage{subfig}
\usepackage{times}
\usepackage{epsfig}
\usepackage{booktabs}
\usepackage{graphicx}
\usepackage{amsmath}
\usepackage{amssymb}
\usepackage{algorithm}
\usepackage{algorithmic}
\usepackage{bm}
\usepackage[inline]{enumitem}
\usepackage{multirow}
\usepackage{xspace}
\usepackage[accsupp]{axessibility}  %
\usepackage{titling}
\usepackage{chapterbib}

\DeclareMathOperator*{\argmax}{arg\,max}

\def\wacvPaperID{999} %

\wacvfinalcopy %

\ifwacvfinal
\fi

\ifwacvfinal
\usepackage[breaklinks=true,bookmarks=false]{hyperref}
\else
\usepackage[pagebackref=true,breaklinks=true,colorlinks,bookmarks=false]{hyperref}
\fi
\usepackage{cleveref} %

\newcommand{\our}{ABAS\xspace}

\newcommand{\ourms}{EMS\xspace}

\begin{document}
\title{Adversarial Branch Architecture Search for Unsupervised Domain Adaptation}

\author{Luca Robbiano\\
Politecnico di Torino\\
Turin, Italy\\
{\tt\small luca.robbiano@polito.it}
\and
Muhammad Rameez Ur Rahman\\
Sapienza University of Rome\\
Rome, Italy\\
{\tt\small rahman@di.uniroma1.it}
\and
Fabio Galasso\\
Sapienza University of Rome\\
Rome, Italy\\
{\tt\small galasso@di.uniroma1.it}
\and
Barbara Caputo\\
Politecnico di Torino, CINI Consortium\\
Turin, Italy\\
{\tt\small barbara.caputo@polito.it}
\and
Fabio Maria Carlucci\\
Huawei Noah’s Ark Lab\\
London, United Kingdom\\
{\tt\small fabiom.carlucci@gmail.com}
}

\predate{}
\postdate{}
\date{}
\pretitle{\begin{center}\Large \bf }
\posttitle{\par\vspace*{12pt}\end{center}}
\maketitle
\ifwacvfinal
\thispagestyle{empty}
\fi

\begin{abstract}
Unsupervised Domain Adaptation (UDA) is a key issue in visual recognition, as it allows to bridge different visual domains enabling robust performances in the real world. To date, all proposed approaches rely on human expertise to manually adapt a given UDA method (e.g. DANN) to a specific backbone architecture (e.g. ResNet).
This dependency on handcrafted designs limits the applicability of a given approach in time, as old methods need to be constantly adapted to novel backbones. 

Existing Neural Architecture Search (NAS) approaches cannot be directly applied to mitigate this issue, as they rely on labels that are not available in the UDA setting. Furthermore, most NAS methods search for full architectures, which precludes the use of pre-trained models, essential in a vast range of UDA settings for reaching SOTA results.

To the  best of our knowledge, no prior work has addressed these aspects in the context of NAS for UDA. Here we tackle both aspects with an  Adversarial Branch Architecture Search for UDA (ABAS): \textbf{i.}\ we address the lack of target labels by a novel data-driven ensemble approach for model selection; and \textbf{ii.}\ 
we search for an auxiliary adversarial branch, attached to a pre-trained backbone, which drives the domain alignment.
We extensively validate ABAS to improve two modern UDA techniques, DANN and ALDA, on three standard visual recognition datasets (Office31, Office-Home and PACS). In all cases, ABAS robustly finds the adversarial branch architectures and parameters which yield best performances. \href{https://github.com/lr94/abas}{https://github.com/lr94/abas}.

\end{abstract}

\section{Introduction}
\label{sec:intro}

\begin{figure*}
    \centering
    \includegraphics[width=0.87\textwidth]{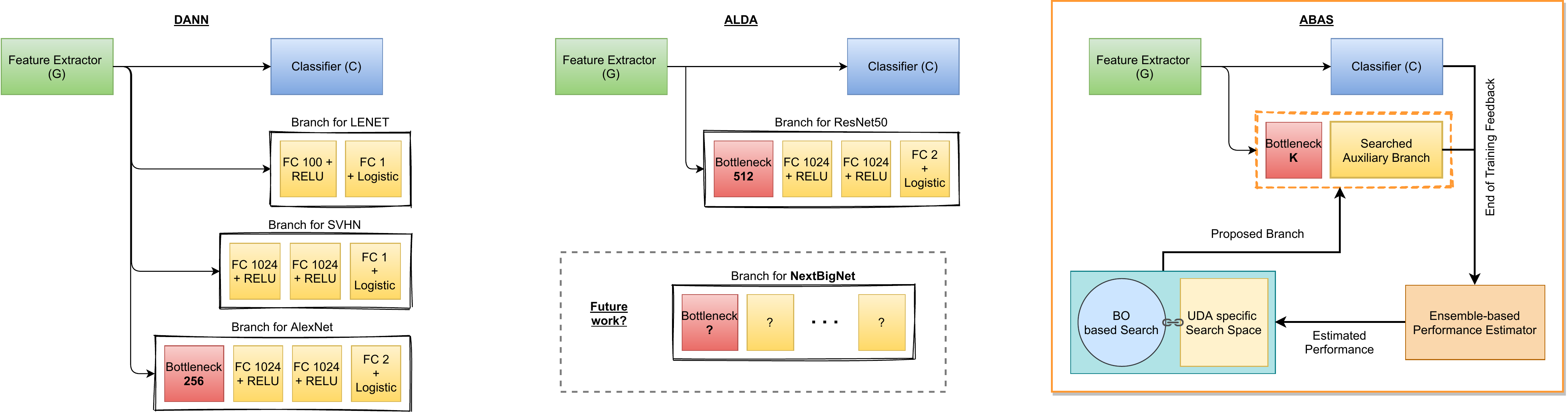}
    \caption{The original DANN \cite{ganin2016domain} paper proposed a specific implementation for each of the considered backbone architectures. Later authors (e.g. ALDA~\cite{chen2020adversarial}) empirically adapted the core idea to different architectures through a process of trial and error. Whenever a new architecture is proposed, this tedious procedure needs to be repeated (note this issue is shared by any other UDA method based on an auxiliary branch). To address this, in \our we propose to automatically search for the optimal architecture of the auxiliary branch in a data driven way. Through \our, it is possible to seamlessly apply DANN to any future architecture without the need for human supervision. What follows is a brief description, more details can be found in Algorithm \ref{alg:main}. At each iteration, the Bayesian Optimization (BO) acquisition function proposes a number of candidate branch architectures. After training, we extract a number of supervised (source) and unsupervised (target) features which are fed into our ensemble-based performance estimator. Its feedback is sent to BO, thus closing the loop.}
    \label{fig:teaser}
    \vspace{-1.0em}
\end{figure*}

Unsupervised Domain Adaptation (UDA) enables the transfer of domain knowledge from a source domain to a target domain, for which possibly few data are available, but no target labels.
Performance has increased steadily in recent years~\cite{chen2020adversarial,li2019cycle,xu2020adversarial,luo2019taking,wu2020dual}, but domain transfer remains constrained to specific setups~\cite{gulrajani2021in,jiang2019neural,lucic2018gans}.
In other words, whenever a new architectural backbone becomes available, optimally adapting existing UDA methods to it requires a manual and slow process of trial-and-error. %

While Neural Architecture Search (NAS) has recently made large progress~\cite{tan2019efficientnet,liu2018darts,tan2019mnasnet,NEURIPS2020_NAGO} in removing the heuristic component of neural architecture design, it does not directly apply to UDA for 
 two main reasons: \textbf{i.}\ NAS requires target labels for model validation when searching for the optimal architecture, which are not available in UDA;  \textbf{ii.}\ the vast majority of UDA methods applied to visual recognition tasks, from classification to detection up to segmentation, require the use of pre-trained (backbone) models to reach state-of-the-art performance on real world datasets~\cite{donahue2014decaf, chen2020adversarial,xu2020adversarial,wu2020dual}. To the best of our knowledge, no research has addressed these essential aspects in the context of NAS for UDA.

\begin{figure}
    \centering
    \includegraphics[width=0.47\textwidth]{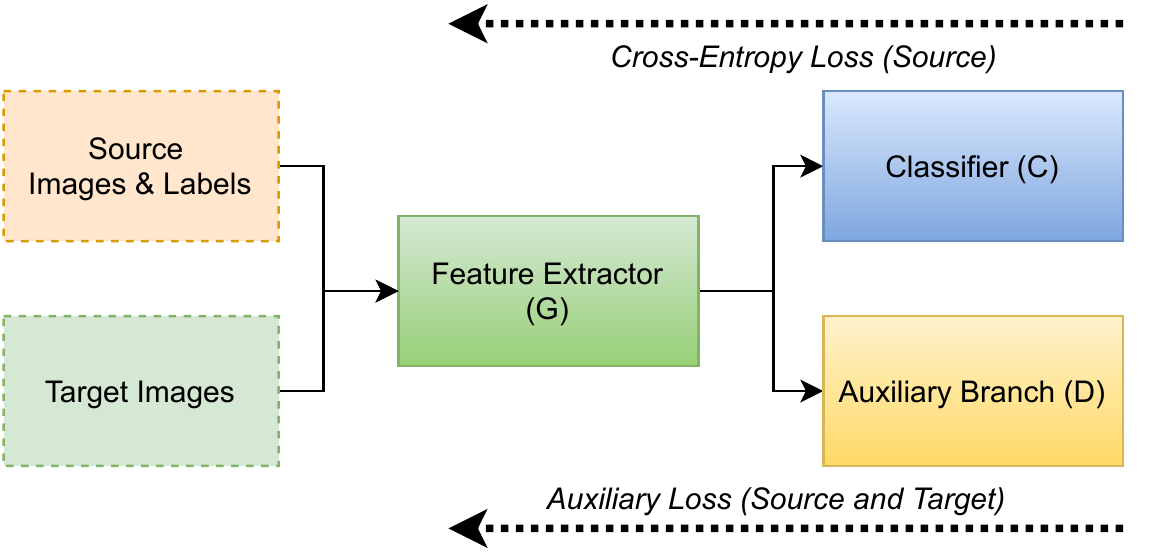}
    \caption{Many UDA methods fall under this general framework. The auxiliary branch $D$ can compute either a domain discriminator \cite{ganin2016domain, chen2020adversarial}, or a self-supervised \cite{carlucci2019domain,sun2019unsupervised} loss.}
    \label{fig:da_branches}
    \vspace{-1.3em}
\end{figure}

We propose a novel Adversarial Branch Architecture Search (\our{} - Fig. \ref{fig:teaser}) for UDA to address both limitations. Our work builds on most recent and robust UDA methods which use \emph{auxiliary adversarial branches}~\cite{ganin2016domain,carlucci2019domain,chen2020adversarial, sun2019unsupervised}: a secondary network (the adversarial branch) is attached to a main backbone (see Fig.~\ref{fig:da_branches} for an illustration). While the main network is trained supervisedly on the source images and labels, the secondary branch reduces the distributional gap between the source and target domains, without target labels, by adversarially making the domains indistinguishable.
The architecture of the auxiliary branch, which plays a huge role in the final accuracy of the method (as shown in Fig.~\ref{fig:random_sampling}),  %
has so far been  hand-crafted.

In \our, as first main contribution, we address the lack of target labels by a novel data-driven Ensemble approach for Model Selection (EMS), which directly applies to target data. Model selection in UDA is challenging and often overlooked. Only very few works have so far focused on it~\cite{you2019towards,nomura2020multi}, but they only consider source features. By contrast, considering the actual target features yields better domain transfers, as we show.

As a second main contribution we propose using a multi-fidelity Bayesian Optimization (BOHB)~\cite{falkner2018bohb} method to search for both the architecture and the hyper-parameters of the auxiliary branch.
\our exploits the backbone architecture and pre-training, which is key to surpass the current state-of-the-art performance.

We conduct an in-depth performance evaluation on a popular UDA technique, DANN~\cite{ganin2016domain}, and on a more recent state-of-the-art method, ALDA \cite{chen2020adversarial}. We test their \our extensions on three computer vision datasets, PACS~\cite{li2017deeper}, Office31~\cite{saenko2010adapting} and Office-Home~\cite{venkateswara2017deep}. 
Our study confirms that the default DANN and ALDA branches 
perform well on
specific backbones and source-target cases, i.e.\ for Office31 and Office-Home, well above random architectures (cf.\ Fig.~\ref{fig:random_sampling} and Sec.~\ref{sec:experiments}). However, they underperform on PACS, where \our also finds the best UDA architecture.

\section{Related work}
\label{sec:rel_work}

\begin{figure*}[t!]
    \centering
    \subfloat{\raisebox{-.5\height}{\includegraphics[width=0.45\textwidth]{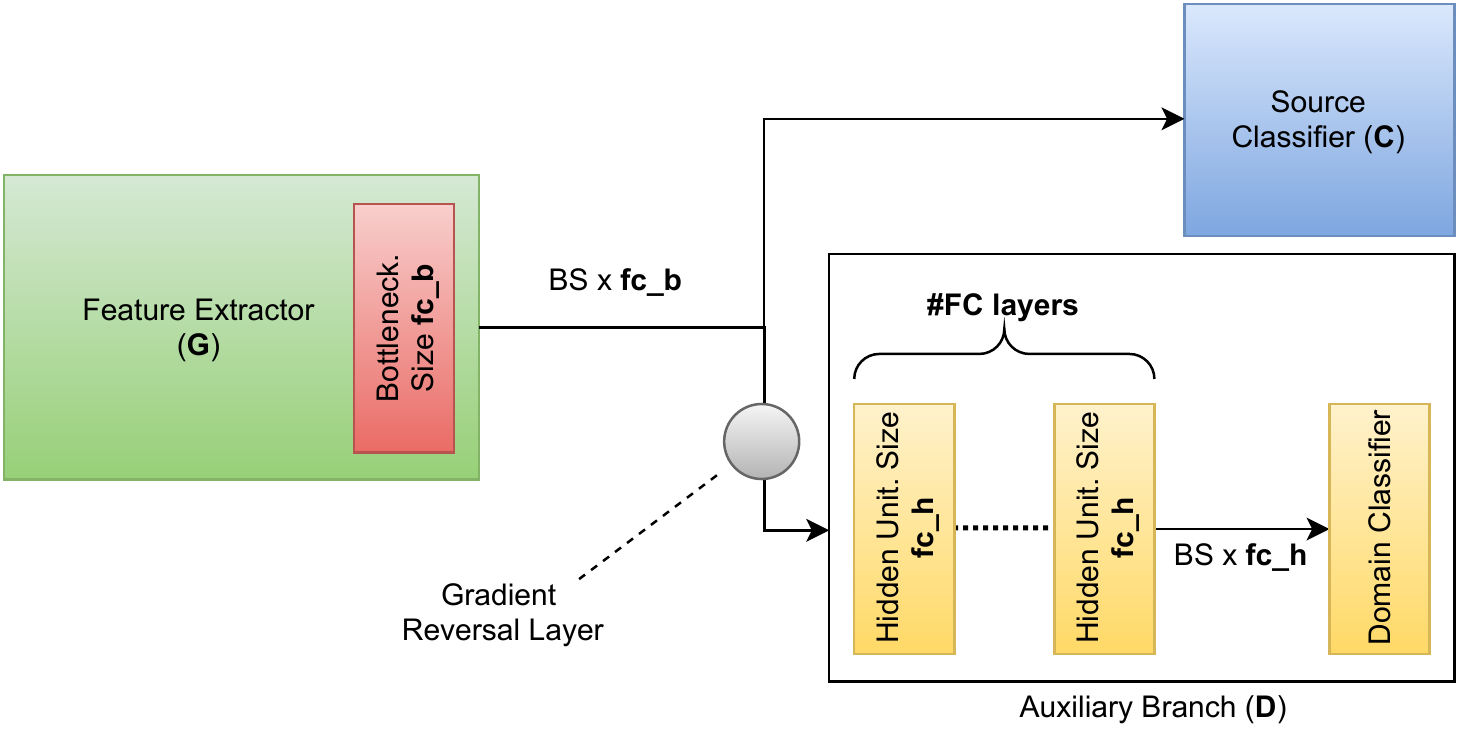}}}
    \hspace{2em}
    \subfloat{
    \begin{tabular}{l|cc}
     \hline
         Hyperparameters & Ranges & Type \\
         \hline
         Domain adaptation weight $\lambda$ & $\left[0,2\right]$ & real \\
         Dropout Probability & $\left[0,1\right]$ & real \\
         \# FC layers & $\left[1,6\right]$ & integer \\ 
         Bottleneck size \textit{fc\_b} & $\left[2^{6},2^{10}\right]$ & integer \\
         Hidden units size \textit{fc\_h} & $\left[2^{6},2^{12}\right]$ & integer \\
         \hline
     \end{tabular}
     }
    \caption{Left: overview of how the \our search space applies to the main architecture (BS is batch size). Right: explicit search space definition. Being low-dimensional it can easily be optimized through BO; on the other hand (as shown in Fig. \ref{fig:param_sensitivity}) each of these parameters has a strong effect on the final accuracy.}
    \label{fig:search_space}
    \vspace{-1.0em}
\end{figure*}

\textbf{Unsupervised Domain Adaptation (UDA).}
Research on UDA has been recently divided into four families of methods \cite{zhao2020review}: discrepancy, adversarial discriminative, adversarial generative and self-supervision based methods.
Adversarial discriminative approaches provide top performance, but they are significantly influenced by the specific architecture of the auxiliary branch, cf.\ Fig.~\ref{fig:random_sampling}: un-optimized architectures may diverge and\slash{}or yield a wide span of accuracies.
We find this strongly motivates research on architecture search for these approaches.

Starting from \cite{ganin2016domain}, UDA has been cast as the problem of learning domain-agnostic features via an auxiliary branch and an adversarial training. The extra branch is attached to the backbone and tasked to learn to discriminate the domains. Then its gradient is reversed to make the backbone features indistinguishable between the source and the target domain. A few recent improvements are notable: \cite{hong2018conditional} considers the cross-covariance between samples;  \cite{NEURIPS2018_ab88b157} takes the classifier prediction into account; \cite{tzeng2015simultaneous} generalizes a similar approach to the semi-supervised setting; \cite{carlucci2019hallucinating} extends the discriminative gradient reversal to the pixel level; and finally \cite{chen2020adversarial} combines the adversarial training with self-training.
All of the above propose specific embodiments of the auxiliary adversarial branch, but offer no guidelines on how to adapt it to different backbones. To the best of our knowledge, this is the first work proposing a data-driven pipeline to learn it.

\textbf{Model Selection for UDA.}
Hyper-parameter and model selection are extremely challenging in the UDA setting, as target labels are not available. In practice, many works simply fix the hyper-parameters across a range of datasets \cite{tzeng2017adversarial,pinheiro2018unsupervised,saito2018maximum,russo2018source}.

A more principled approach based on importance-weighted cross-validation (IWCV) was first proposed by \cite{sugiyama2007covariate}: the more the source sample is similar to the target, the higher the importance. %
The IWCV estimator is unbiased, but has unbounded variance; Deep Embedded Validation~\cite{you2019towards}, was proposed to reduce the variance and adapt IWCV to deep features. Similarly, \cite{nomura2020multi} proposes a reduced-variance IWCV estimator for hyper-parameter optimization in the context of multi-source DA.
The current literature focuses
on the source, only implicitly taking target features into consideration. By contrast, we propose a performance estimator which  mostly uses the actual target features (see Sec.~\ref{subsec:model_selection}).

\textbf{Neural Architecture Search (NAS) and UDA-NAS.}
There is a wealth of literature on NAS~\cite{DBLP:conf/iclr/ZophL17} leveraging diverse search strategies: Bayesian Optimization (BO), evolutionary, reinforcement learning, and gradient-based, as recently surveyed in \cite{elsken2019neural,ren2020comprehensive}.
BO is one of the most efficient strategies for sample-based hyper-parameter tuning \cite{snoek2012practical}, and has been recently applied to NAS \cite{white2019bananas, ying2019bench}. Multi-fidelity BO (BOHB) \cite{falkner2018bohb}, which further reduces the search cost by performing partial evaluations, has also been successfully applied to NAS \cite{zela2018towards,NEURIPS2020_NAGO}.
\our exploits pre-trained backbones and efficiently fine-tunes the architectures, enabling us to use a more accurate sample-based technique. In particular, we adopt BOHB for its sota performance.

A recent pre-print~\cite{li2020network} has proposed a DARTS-like~\cite{liu2018darts} method for UDA. While that shares similar motivations with \our, it is unable to leverage pre-trained models and as such it can only be applied to simple low-resolution datasets. Furthermore model selection is performed by a simple validation on the source domain, which we show to be insufficient in the general case.

\section{Method}
\label{sec:method}
We propose a data-driven pipeline to search for the optimal architecture of the auxiliary branch in the context of unsupervised domain adaptation. 
As mentioned in section \ref{sec:rel_work}, a number of UDA approaches require the use of an auxiliary branch: 
while we explicitly explore how this idea applies to works based on domain adversarial learning, it is trivially extendable to other methods (see section \ref{sec:discussion}). Here we apply \our to 1) the first UDA adversarial approach, DANN \cite{ganin2016domain} and to 2) the current state-of-the-art variation, ALDA \cite{chen2020adversarial}. In both cases, the auxiliary branch is adversarially trained to reduce the distributional gap at the feature level.

\subsection{Background}
\label{subsec:back}

Due to space constraints, we provide here only a mostly intuitive overview and refer to the referenced papers for a formal presentation of background methods.

\textbf{Adversarial branches for domain adaptation}
Let us define a dataset $\bm{X}_s=\{\bm{x}^i_s,y^i_s\}_{i=0}^{N_s}$
drawn from a labeled source domain $\mathcal{S}$, and a dataset $\bm{X}_t=\{\bm{x}^j_t\}_{j=0}^{N_t}$ from a different 
unlabeled target domain $\mathcal{T}$, sharing the same set of categories.  
Our goal is to maximize the classification accuracy on $\bm{X}_t$ while training on $\bm{X}_s$.
As it can be seen in Figure \ref{fig:da_branches}, the overall architecture consists of 3 components: the feature extractor $G$, the labeled classifier $C$ and the adversarial branch $D$. 
The basic intuition behind discriminative adversarial approaches for domain adaptation is that a domain classifier is trained to distinguish between source and target samples; as its gradient is reversed (Fig. \ref{fig:search_space}), the features in $G$ are trained so that source and target have similar representations.
The model is trained by optimizing the following objective function:
\begin{equation}
\min_{{G}, {C}, {D}} ~~ \mathcal{L}_{CE}(C(G(x_s)), y_{s}) + \lambda \mathcal{L}_{Adv}(D(G(x_{s} \cup x_t))
\label{eq:advloss}
\end{equation}
Where $\mathcal{L}_{CE}$ is the standard cross-entropy loss, computed on source samples only, and $\mathcal{L}_{Adv}$ is the adversarial loss trained to reduce the domain gap. In the case of DANN \cite{ganin2016domain},
\begin{equation}
    \mathcal{L}_{Adv}(D(G(x_i)), d_i) = d_i \frac{1}{D(G(x_i))} + (1-d_i)\frac{1}{1-D(G(x_i))}
    \label{eq:dann}
\end{equation}
With $d_i$ being a binary variable (domain label) indicating whether the sample $i$ belongs to the source or to the target.

ALDA \cite{chen2020adversarial}, is a modern DANN variant, currently state-of-the-art, which combines domain adversarial training with self-training. While its overall loss contains multiple terms, we focus here only on the adversarial loss: %
\begin{equation} \label{eq:ALDA}
    \begin{aligned}
        \mathcal{L}_{Adv}(x_s, y_s, x_t) =  \\
        \mathcal{L}_{BCE}(c^{(x_s)},y_s) + \mathcal{L}_{BCE}(c^{(x_t)}, u^{(\hat{y}_t)})
    \end{aligned}
\end{equation}
$L_{BCE}$ is the Binary Cross-Entropy loss, $c^{(x)}$ is the corrected pseudo-label vector generated by multiplying the confusion matrix and the pseudo-label vector $\hat{y}$, and $u^{(\hat{y}_t)}$ is the opposite distribution of $\hat{y}_t$.
Intuitively, the authors propose to use a discriminator branch to predict the confusion matrix instead of the domain probabilities. The confusion matrix is then used to correct the pseudo-label vector obtained by feeding target samples into the network, which is then used as the target label for the training process. The discriminator is trained to produce different corrected vectors for each domain, while the feature extractor learns to fool it.
\our will be used  to search for the best performing architecture for $D$ in both DANN and ALDA.

\paragraph{Bayesian Optimization for AutoML}
Several optimization strategies could be used to guide the search, but our search space is compact and has low dimensionality (cf.\ Sec.~\ref{subsec:search_space}), which makes it particularly suitable for Bayesian Optimization (BO).
For efficiency, we adopt BOHB, a multi-fidelity combination of BO and Hyperband \cite{li2017hyperband}. While single fidelity BO evaluates all samples with full budget, BOHB resorts to partial evaluations with smaller-than-full budget, excluding bad configurations early in the search process and reserving computational resources for the most promising configurations. So, given the same time budget, it evaluates many more configurations and it %
achieves faster optimization than competing methods.
Details of the BOHB's role in \our are reported in Sec. \ref{subsec:overall_method} %

\subsection{\our Search space}
\label{subsec:search_space}

UDA methods for visual object categorization require the use of a pre-trained model (typically on ImageNet) to be competitive. %
Hence, the architecture of $G$ and $C$ are usually fixed as a subset of the backbone: in practice the only choice being made is where $G$ ends and $C$ begins (see Fig. \ref{fig:da_branches}). 
Our search space (Fig. \ref{fig:search_space}) is a combination of hyper and architectural parameters: the domain loss weight $\lambda$ (eq. \ref{eq:advloss}) and the dropout probability of the fully connected layers in $D$ are hyper-parameters, while the number of fully connected layers (\#fc layers), size of the hidden units (\textit{fc\_h}) and of the bottleneck (\textit{fc\_b}) fully characterize the width and the height of the auxiliary branch $D$.
Giving the bottleneck layer a separate width allows to control the overall capacity of the auxiliary branch by limiting how much information it is fed.

\subsection{Ensemble-based Model Selection (\ourms)}
\label{subsec:model_selection}

\begin{figure*}[h]
\centering
\subfloat{\includegraphics[width = 0.25\textwidth]{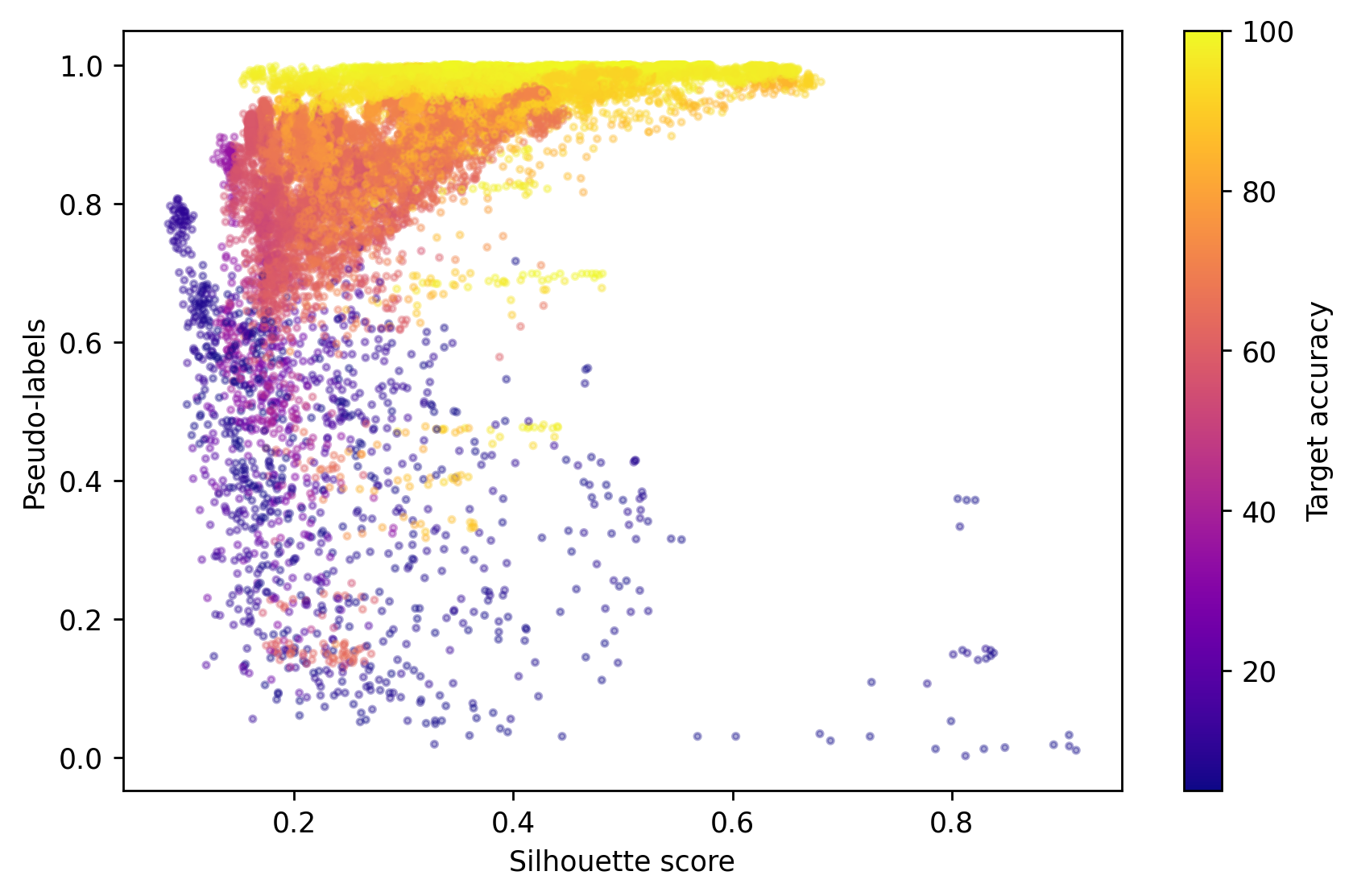}} 
\subfloat{\includegraphics[width = 0.25\textwidth]{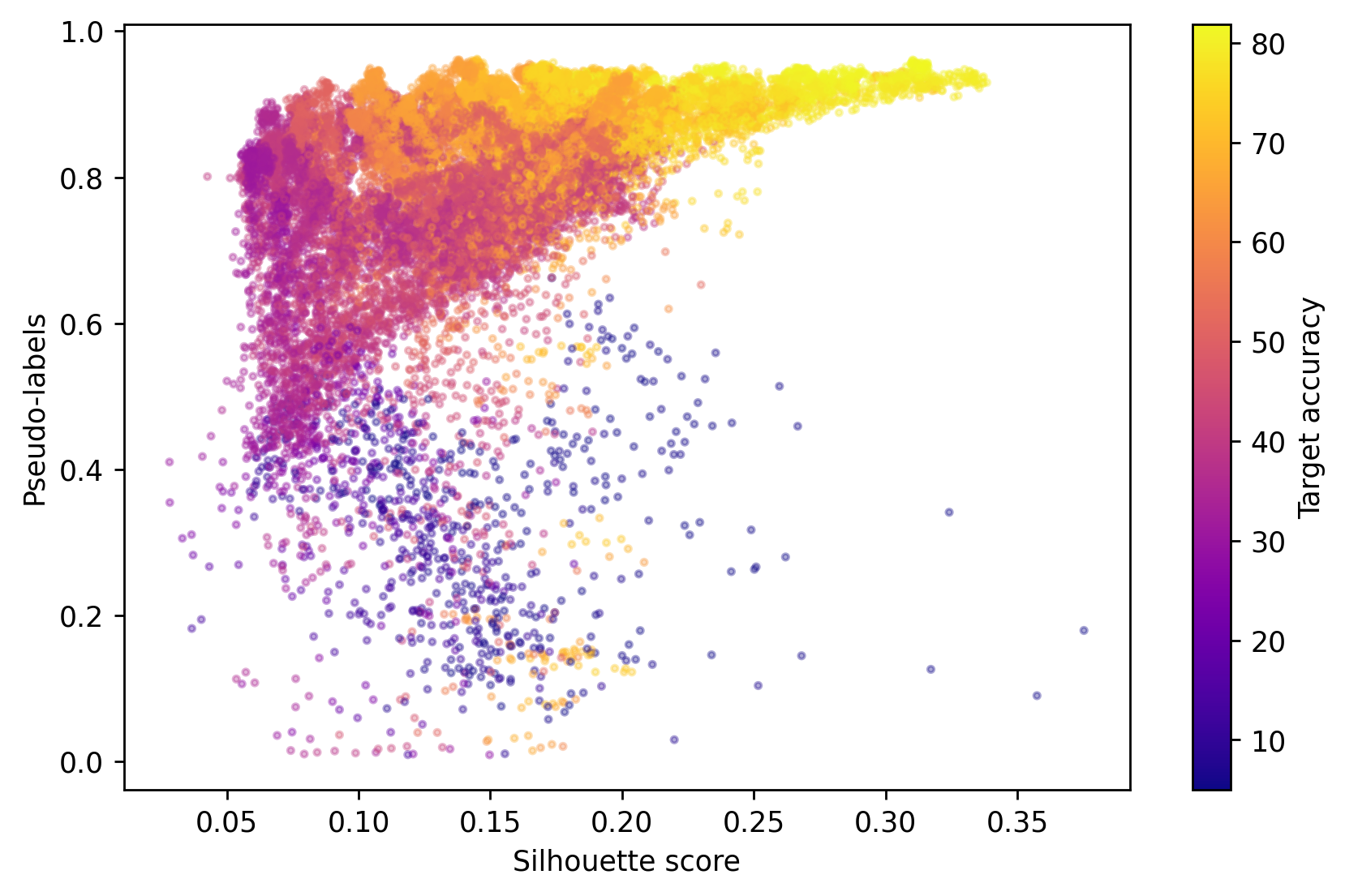}}
\subfloat{\includegraphics[width = 0.25\textwidth]{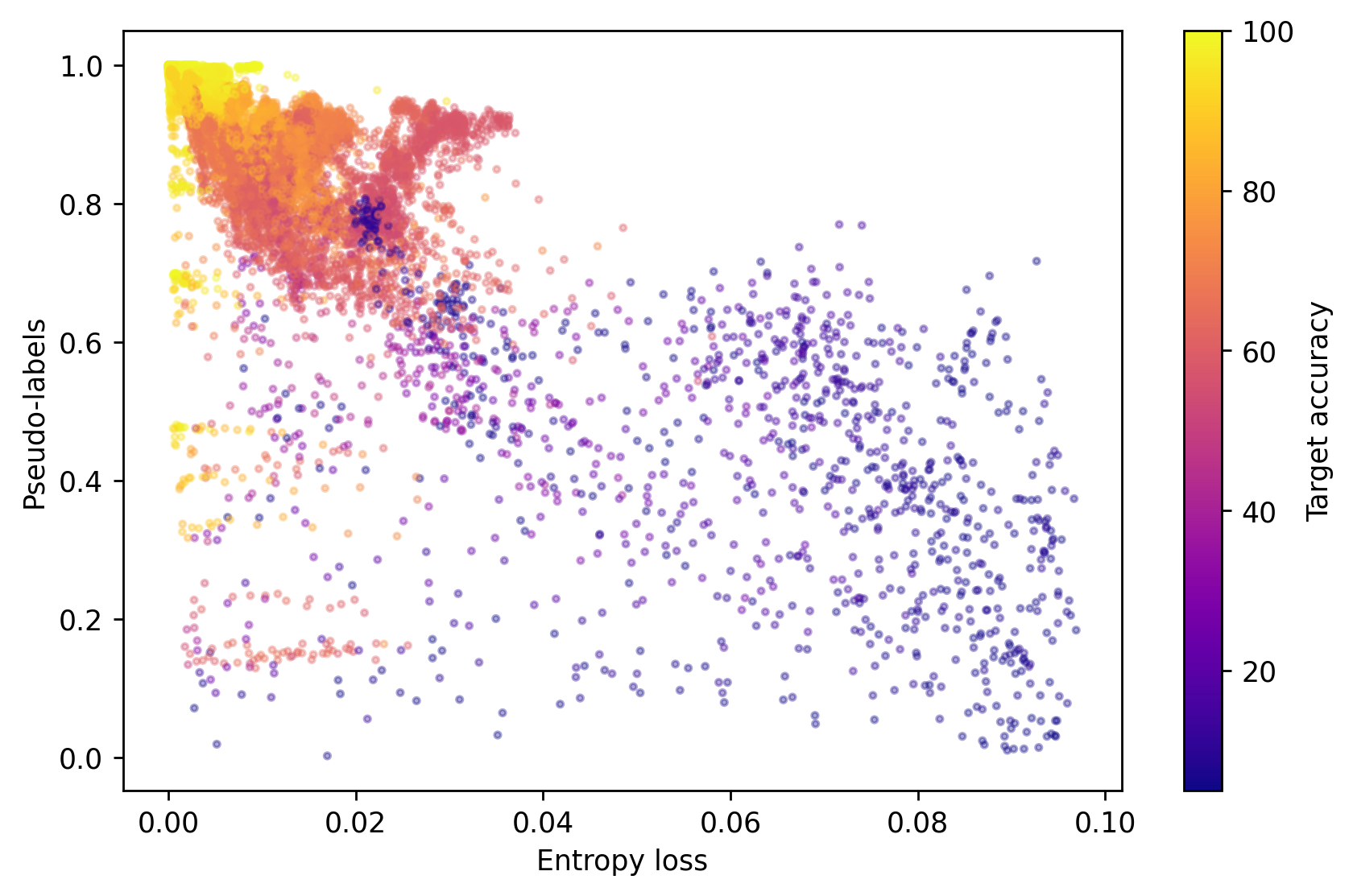}}
\subfloat{\includegraphics[width = 0.25\textwidth]{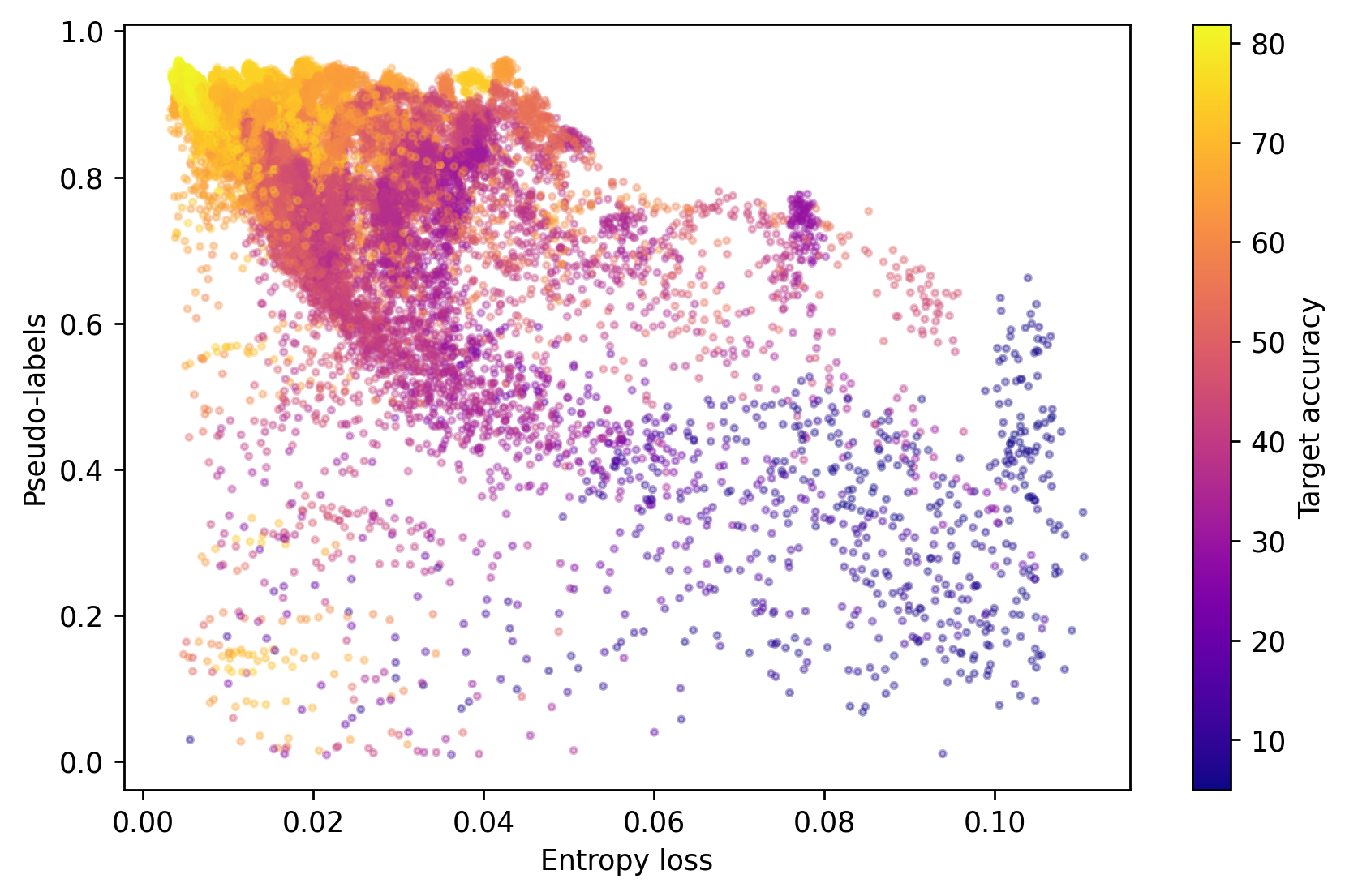}}
\caption{Different model configurations trained on Office31 (1st and 3rd) and Office-Home (2nd and 4th). The x and y axes show two different unsupervised validation metrics we use in \ourms, while the color represents target accuracy.
Clearly, by only using a single metric it is extremely hard to uniquely predict the performance, but the combined use of two metrics makes it easier to distinguish between good and bad models.
This shows how different metrics can be complementary and employed to build a better estimator of model performance. The similarity between pairs of plots on different settings (1\&3 and 2\&4) suggests that an estimator built on one dataset can be successfully transferred on another. Further plots can be found in the Supplementary.}
\label{fig:bimetric_o31_oh}
\vspace{-1.3em}
\end{figure*}

Model selection in the context of UDA is extremely challenging, as target labels are not available. To overcome this, we propose using %
an ensemble of weakly correlating predictors. Experimentally, we show how a linear regressor over these metrics, trained on one dataset, is capable of capturing the ranking between models on a \textit{different} unseen dataset, thus highlighting the generalization power of this approach.

A description of the considered metrics follows ([S] or [T] implies on which domain the metric is computed).
\begin {enumerate*}
\item \textbf{Entropy}~[T] \cite{grandvalet2005semi} minimization is commonly used in UDA methods as a regularizer for unlabeled samples \cite{roy2019unsupervised}; intuitively a low entropy corresponds to well separated classes. Mathematically the target entropy is defined as $H(C(G(X_T))) = - \frac{1}{N_t} \sum_{x^i \in X_T}C(G(x^i)) log(C(G(x^i)))$.
\item \textbf{Diversity}~[T], recently proposed by \cite{wu2020entropy}, measures the diversity of class predictions in a batch. It is defined as the entropy of the predicted category distributions, $H(\hat{q}(\mathcal{T}))$.
\item \textbf{Silhouette and Calinski-Harabasz score}~[T] \cite{rousseeuw1987silhouettes, calinski1974dendrite}: the most confident label is assigned to the target samples, which are then clustered with K-Means using the predicted class centroids. The Silhouette and Calinski score can then be used to estimate how well separated are the different classes.
\item \textbf{Source (weighted) loss}~[S]: the cross-entropy loss can be computed on the labeled source samples to estimate the performance of the predictor. Additionally it is possible to weight the samples according to their similarity to the target, as in Importance Weighted Validation \cite{sugiyama2007covariate,you2019towards}. 
We initially experimented with the source-weighted accuracy in \ourms, but due to a number of practical drawbacks we decided to adopt the vanilla source accuracy instead. Specifically, the performance of Importance Weighting methods heavily depends on how well the regression model estimates the density ratio, which is not trivial to successfully implement; secondly, it requires a separate source validation set (not always available on small settings such as Office31).
\item \textbf{Time consistent pseudo-labels}~[T]: inspired by \cite{zhou2020time}, we propose to use the mode of all model predictions during training (one prediction per epoch) as target pseudo-labels, as %
the label predicted more often is likely to be the true one.
Once training has ended, we compute these pseudo-labels and use them to retroactively estimate the target accuracy at each epoch. %
\end{enumerate*}

Note how all of these metrics, with the exception of the source accuracy, are computed on the target domain. 
These metrics are all relatively cheap to compute and have good correlation on their own with the target accuracy, but as we show in Fig. \ref{fig:bimetric_o31_oh} and in the Supplementary, they are all individually vulnerable to degenerate cases. To overcome this, we learn a simple least-squares linear regressor to predict the target accuracy. Since it is not possible to use the target labels of interest we instead learn the regressor on one experimental setting and then use it on a different one. It is worth mentioning that although the output of the regressor cannot be used to estimate the \textit{accuracy} of a model on an unseen dataset, its output has a strong correlation with the ground-truth ranking, as we show in section \ref{subsec:prelim_analysis}. We use this predictor to not only return  the best sample, but also the best snapshot during training.

\subsection{Auxiliary branch optimization}
\label{subsec:overall_method}
A full overview can be found in Fig. \ref{fig:teaser} and Algorithm \ref{alg:main}: \our combines BOHB \cite{falkner2018bohb} and an adversarial method of choice (sec. \ref{subsec:back}), with our UDA specific search space and model selection strategy.
Given a fixed budget, \our performs a number of rounds, alternating between sampling and evaluating. At each step, we sample the Bayesian acquisition function $\alpha(\boldsymbol{\Theta} \vert D)$ for $B$ different configurations $\{\boldsymbol{\Theta}^j_{t}\}_{j=1}^B$. The configurations are used to build auxiliary branches for method $Q$ and the resulting network is trained on the target setting. After training, supervised (from the source) and unsupervised (from the target) features are collated for our model selection module, as described in section \ref{subsec:model_selection}. The ensemble predictor finally gives feedback to the BO process. This procedure is repeated for a given number of rounds and in the end the best performing model is returned.
The pipeline itself is relatively simple and, as shown in the next section, capable of significantly improving over existing sota methods. Finally, while we decided to use BOHB in this particular implementation, it would be feasible to use a different search strategy.

\begin{algorithm}
[tb]
   \caption{BO auxiliary branch optimization}
   \label{alg:main}
\begin{algorithmic}[1]
   \STATE {\bfseries Input:} Domain adaptation method $Q$, Performance estimator $E$, BO surrogate model $p(f\vert \boldsymbol{\Theta}, D)$ and acquisition function $\alpha(\boldsymbol{\Theta} \vert D)$
   \FOR{$t=1$ {\bfseries to} $T$}
   \STATE Recommend $\{\boldsymbol{\Theta}^j_{t}\}_{j=1}^B=
   \argmax 
   \alpha_{t-1}(\boldsymbol{\Theta} \vert D)$
        \FOR{$j=1$ {\bfseries to} $B$ {\bfseries in parallel}}
        \STATE Build the auxiliary branch and evaluate $E(Q(\boldsymbol{\Theta}^j_t))$ to obtain its corresponding performance metric $f^j_t$ 
        \ENDFOR
   \STATE Update $D$ and thus $p(f\vert \boldsymbol{\Theta}, D)$ with $\{\boldsymbol{\Theta}^j_{t}, f^j_t\}_{j=1}^B$
   \ENDFOR
   \RETURN The best model according to $E(Q(\boldsymbol{\Theta}^*))$
\end{algorithmic}
\end{algorithm}

\section{Experiments}
\label{sec:experiments}
\begin{figure*}[t]
    \centering
    \subfloat{\includegraphics[width = 0.3\textwidth]{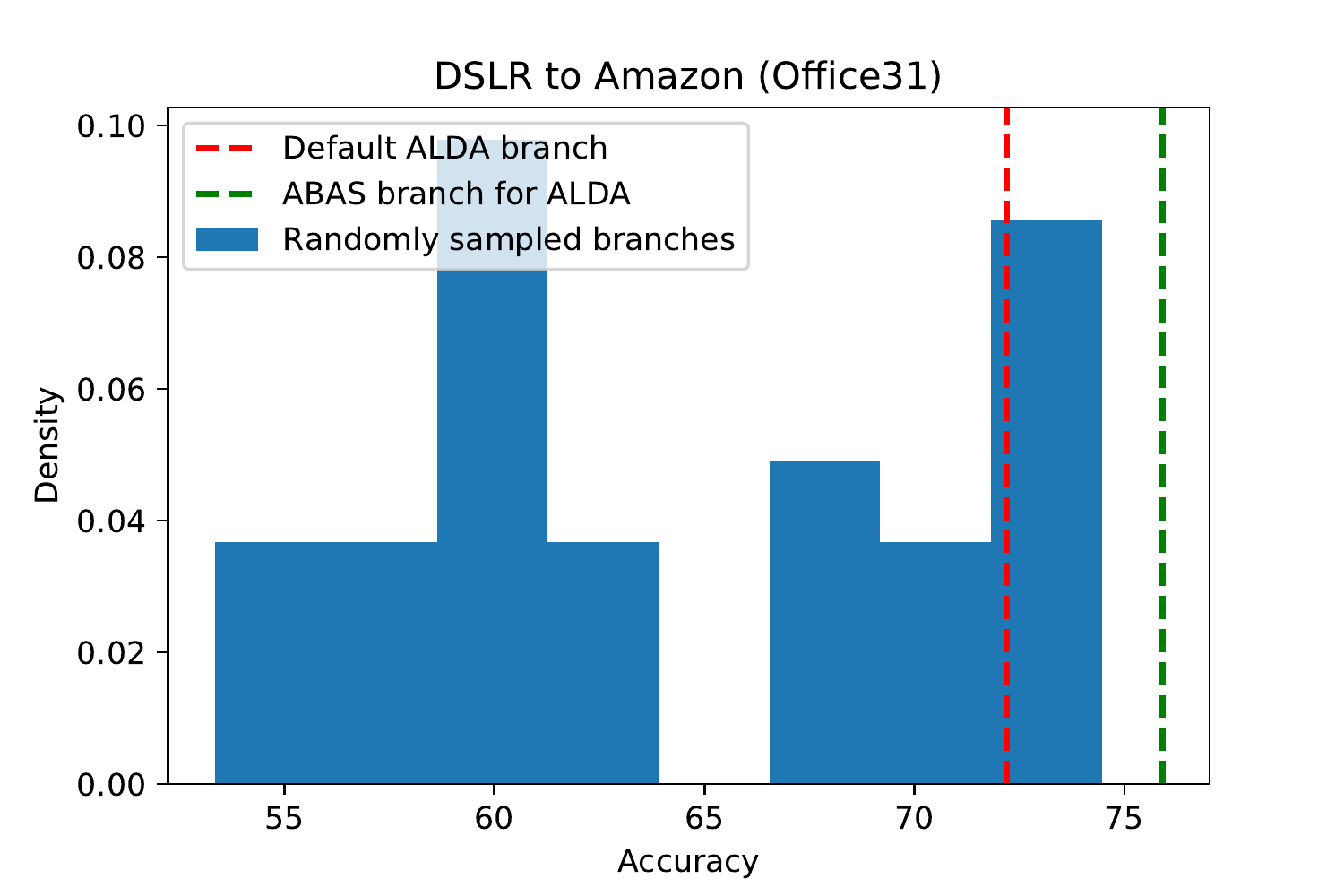}} 
    \subfloat{\includegraphics[width = 0.3\textwidth]{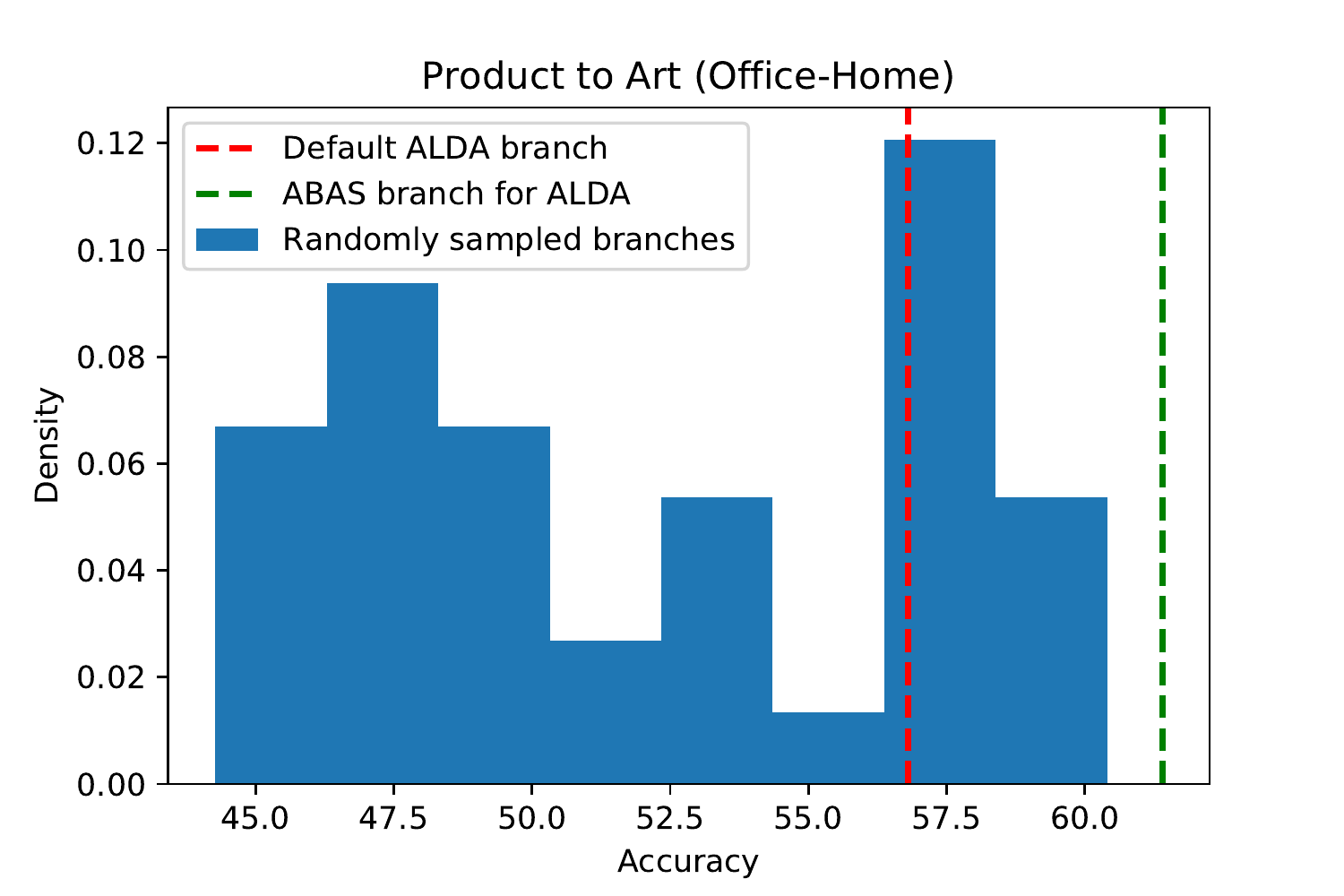}}
    \subfloat{\includegraphics[width = 0.3\textwidth]{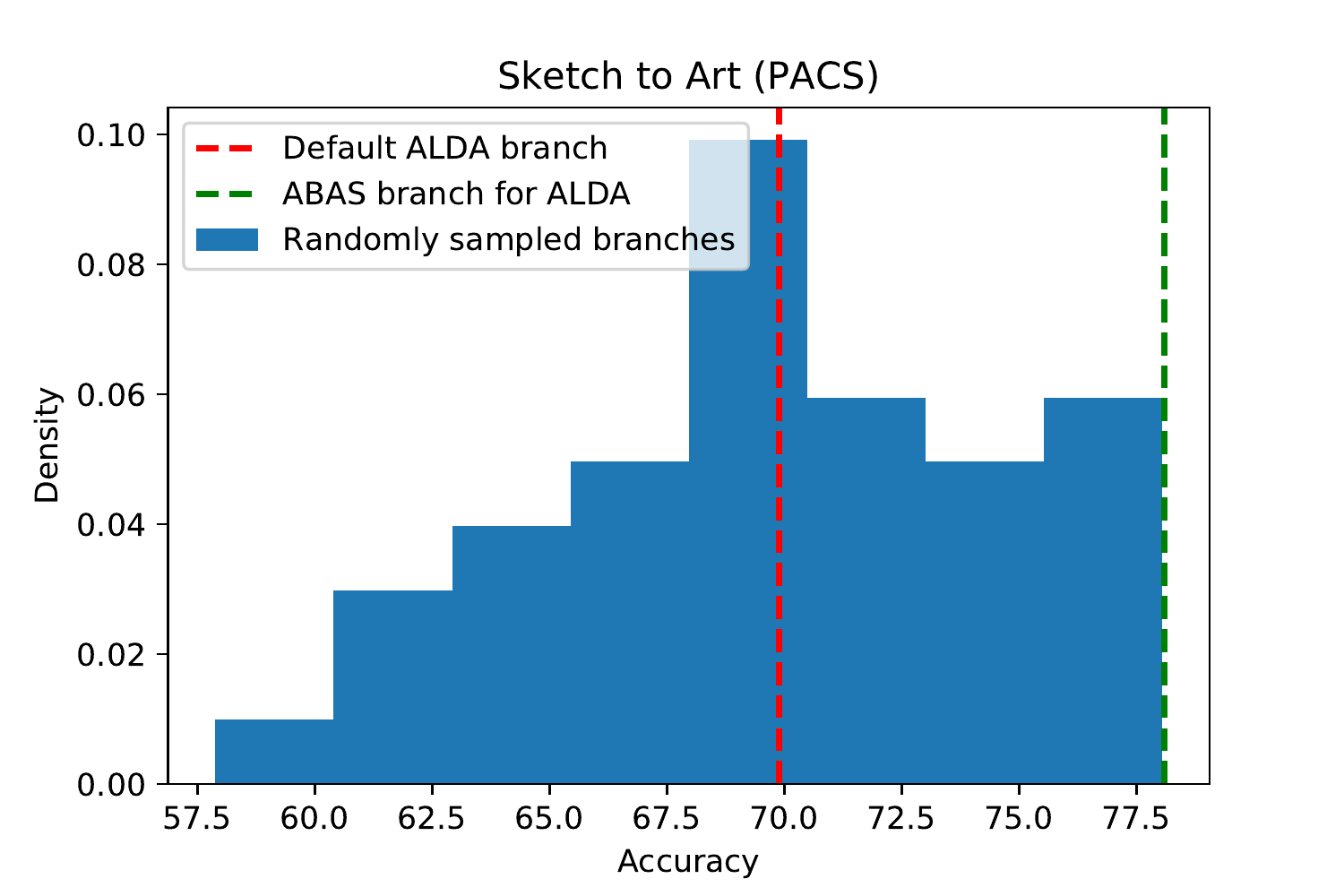}}
    \caption{Accuracy histograms obtained by randomly sampling $40$ architectures for ALDA's~\cite{chen2020adversarial} auxiliary branch  and training them on three settings ($10\%$ didn't converge and were excluded). Contrary to many NAS search spaces \cite{Yang2020NASEFH}, ours is large enough to offer a significant impact on final performance (around $20\%$) on all settings. Equally important, while the default ALDA branch performs well on Office31 and Office-Home (the datasets it was tested on), it performs significantly worse on PACS. \our, on the other hand, can adapt itself to all settings, consistently outperforming the random sampling baseline. See sec.~\ref{subsec:prelim_analysis} for full details.
    }
    \label{fig:random_sampling}
    \vspace{-1.3em}
\end{figure*}

\textbf{Datasets}. We evaluate our method on three publicly-available multi-domain datasets.
The \textbf{Office31} dataset \cite{saenko2010adapting}, composed of $4,652$ images, is a standard benchmark for domain adaptation. The images depict objects belonging to $31$ categories and taken from three different domains: Amazon (A), DSLR (D), and Webcam (W). Webcam and DSLR pictures were manually captured in the same office environment with a consumer webcam and a reflex camera.
Amazon samples are images of products downloaded from amazon.com.
The \textbf{Office-Home} dataset \cite{venkateswara2017deep} contains objects from $65$ categories and a total of around $15,500$ images from four domains: Art, Clipart, Product, and Real-World. Like Office31's Amazon domain, the Product domain features product pictures taken from websites. Real-world images were captured with a regular digital camera.
\textbf{PACS} \cite{li2017deeper} is composed of $9,991$ images from four domains, each %
containing objects from $7$ categories. While PACS is commonly used for domain generalization, we employ it in our single-source setting to explore a diverse test-bed.

\textbf{Baseline methods}
To highlight the improvements brought by our \our pipeline, we apply it to two representative UDA methods:
DANN \cite{ganin2016domain}, the original domain adversarial approach,  to this day and base on which all methods of this family build on, and ALDA \cite{chen2020adversarial} the current state-of-art among discriminative adversarial approaches.
\textbf{Training protocol}
\our is based on an outer BOHB loop alternated with an inner loop in which different networks are trained.
For BOHB, we set $\eta = 3$, minimum budget $2000$ iterations, maximum budget $6000$ iterations and run it for $24$ iterations. Intuitively, the most promising solutions evaluated at a lower budget will be re-run at a higher budget. For each sample, BOHB indicates a computational budget and an architectural configuration, which are used to build the appropriate auxiliary branch in the context of the adversarial method of choice.

In all experiments we use a ResNet50 \cite{he2016deep} backbone pretrained on ImageNet; the adversarial branch (built according to each specific sample, see sec. \ref{subsec:search_space}) is located immediately after a bottleneck layer preceding the final classification layer (Fig. \ref{fig:search_space}). Each model is trained end to end, using the same training code and protocol as \cite{chen2020adversarial}: on all settings, we use SGD optimizer, batch size $36$, momentum $0.9$, weight decay $0.0005$ and initial learning rate of $0.001$. We adjust the learning rate 
by $ \mu_p= \frac{\mu_0}{(1 + \alpha \cdot p)^{\beta}}$, where $p$ is the training progress linearly changing from $0$ to $1$, $\mu_0 = 0.001$, $\alpha = 10$ and $\beta = 0.75$, following  \cite{ganin2016domain}, the weight of the gradient reversal layer is gradually increased from 0 to 1: $\rho = \frac{2}{(1 + exp(-\gamma \cdot p))} - 1 $ where $\gamma$ is set to 10 for all experiments. %
Since the bottleneck and the auxiliary branch are trained from scratch, we use a higher learning rate ($10 \mu_p$) and weight decay ($0.001$) for those parts of the network.
At each iteration, a source and target batch are fed into the backbone. The cross-entropy loss is computed on the source, while both source and target samples contribute to the adversarial branch loss (DANN or ALDA loss, depending on the setting). At the end of training, both source and target features are then fed into the performance estimator (as detailed in section \ref{subsec:model_selection}) to provide feedback to the Bayesian Optimization process.

To train our regressors, we randomly sampled $200$ configurations on each dataset and trained them while collecting $100$ snapshots for each run (total of $20,000$ data points). At each snapshot we collected the metrics described in sec. \ref{subsec:search_space} thus building a dataset to train our regressor on; note that we only use the pseudo-label metric to assess the best snapshot in a single run, not to compare across runs. All input features were standardized to zero mean and unit standard deviation. For the regressor we used a linear least-squares model as implemented in Scikit-Learn \cite{scikit-learn}. 
On Office31 we used a regressor trained on Office-Home and vice-versa. The PACS regressor was trained on Office31.

\begin{figure*}[h]
    \centering
    \subfloat{\includegraphics[width = 0.255\textwidth]{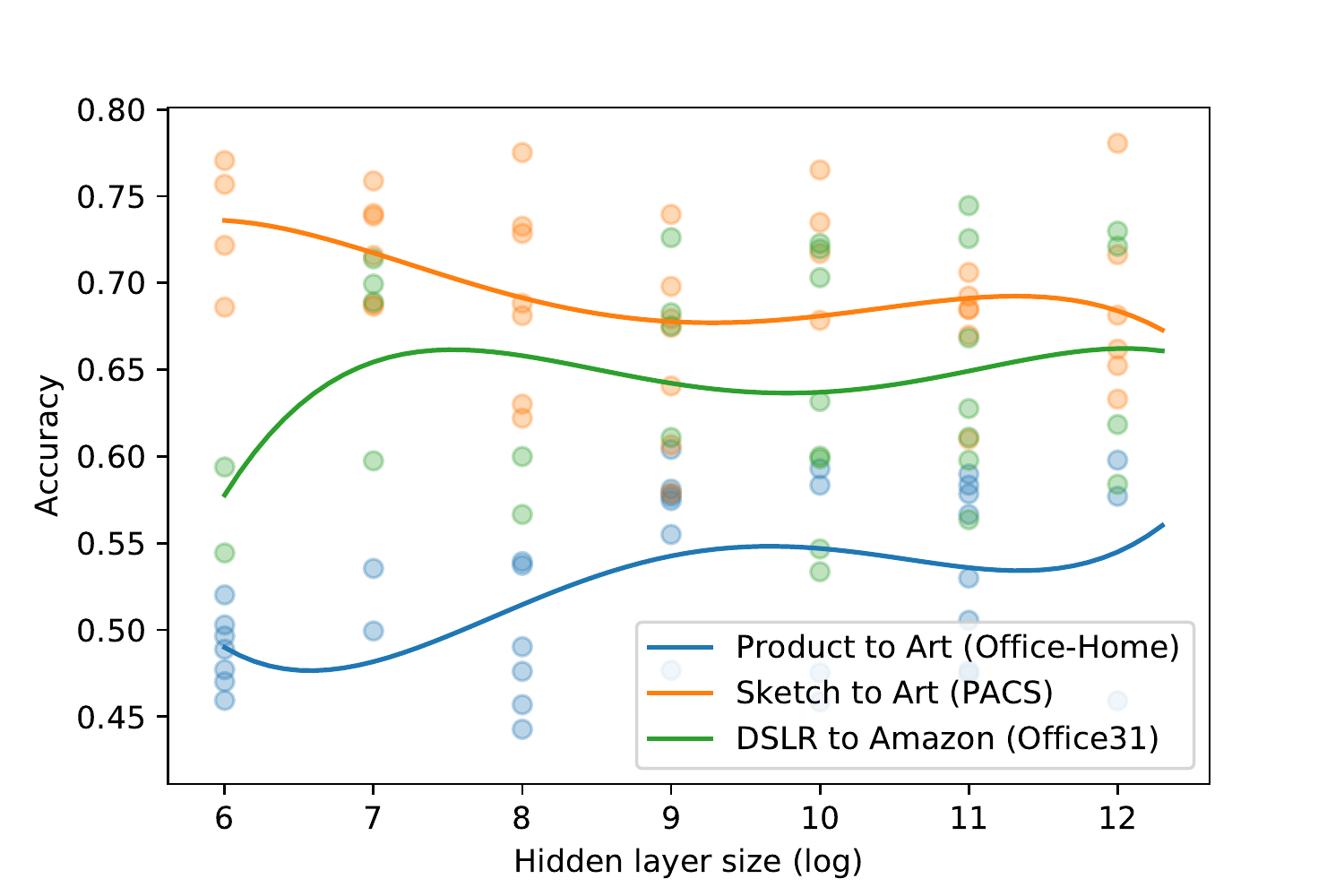}} 
    \subfloat{\includegraphics[width = 0.255\textwidth]{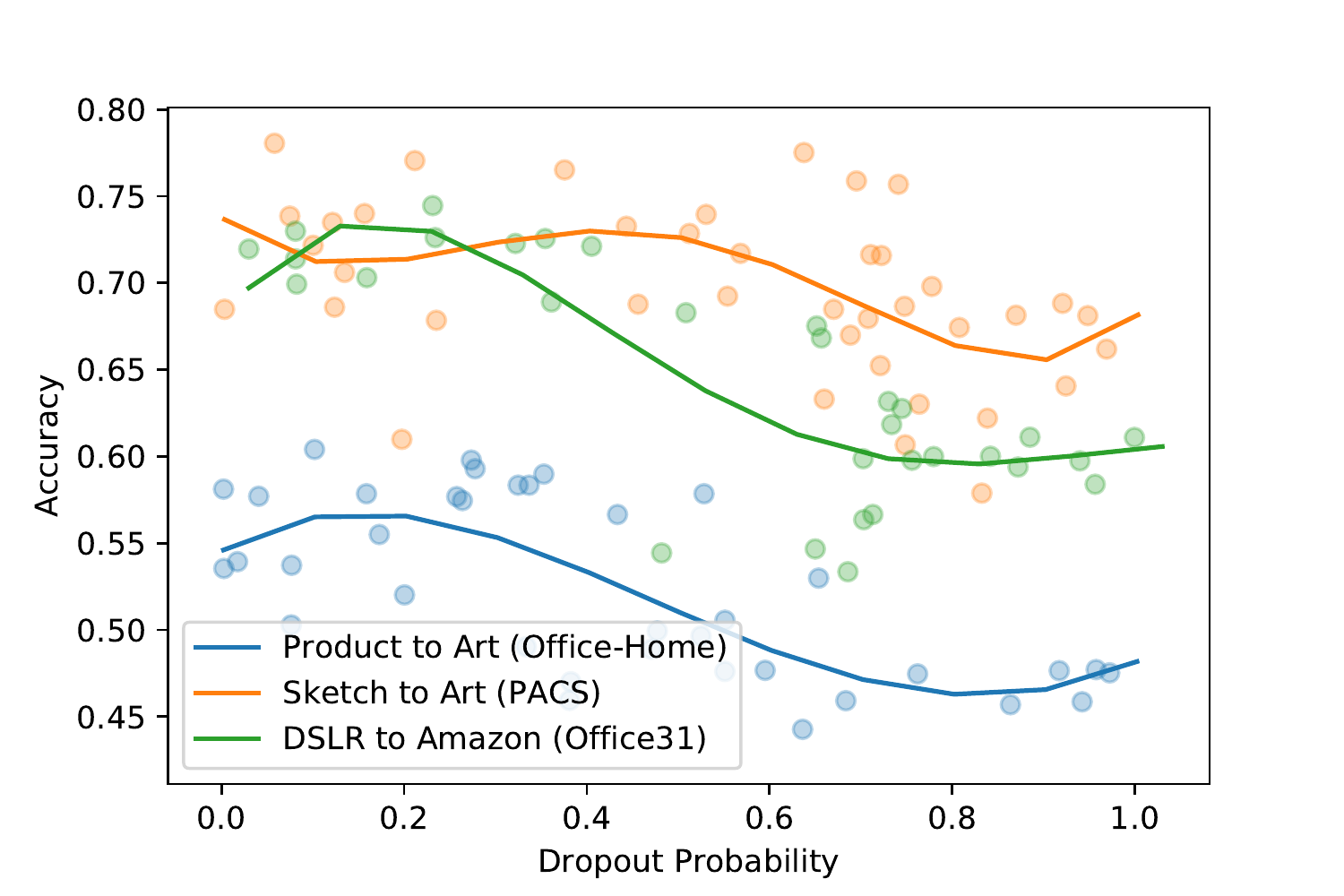}}
    \subfloat{\includegraphics[width = 0.255\textwidth]{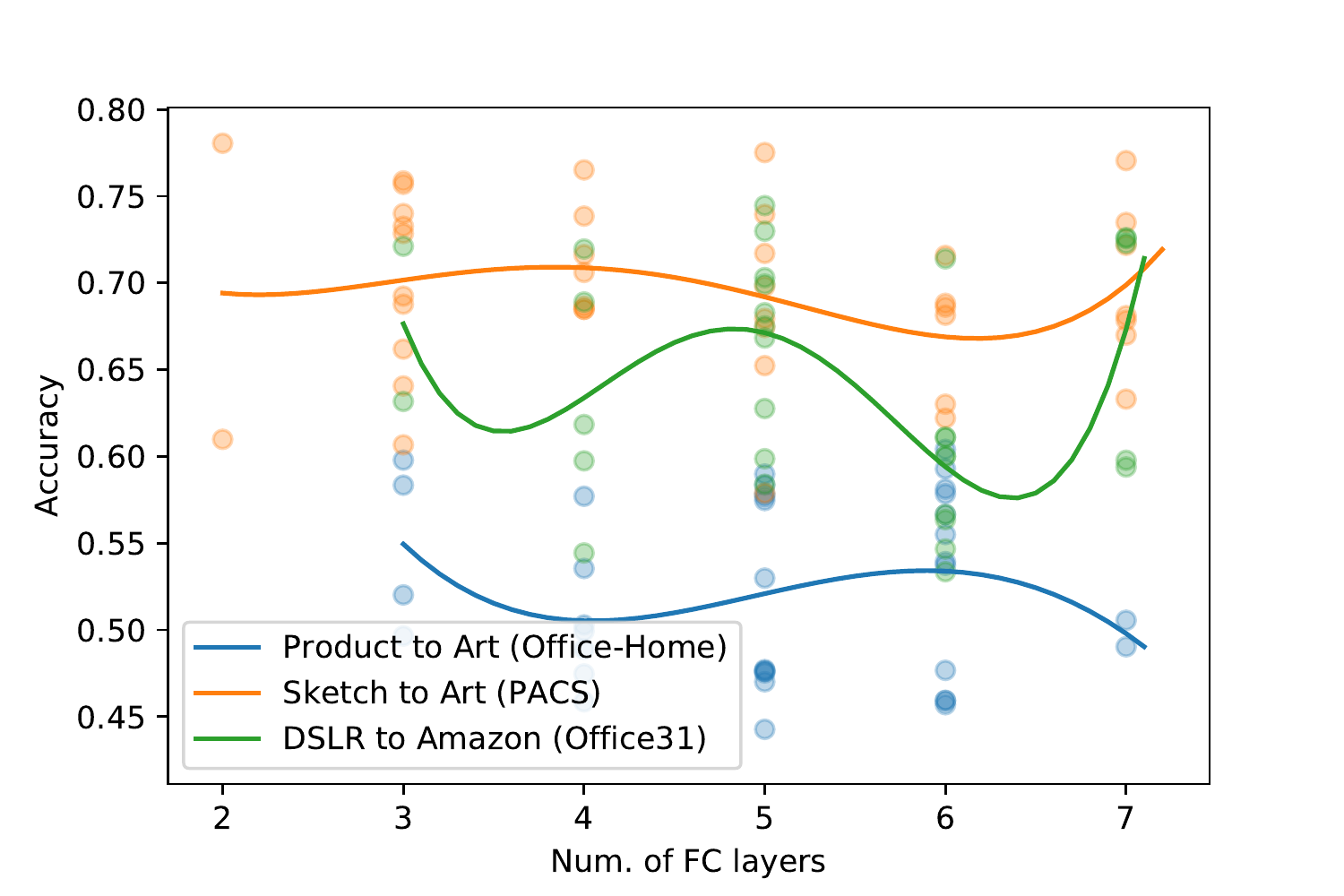}}
    \subfloat{\includegraphics[width = 0.255\textwidth]{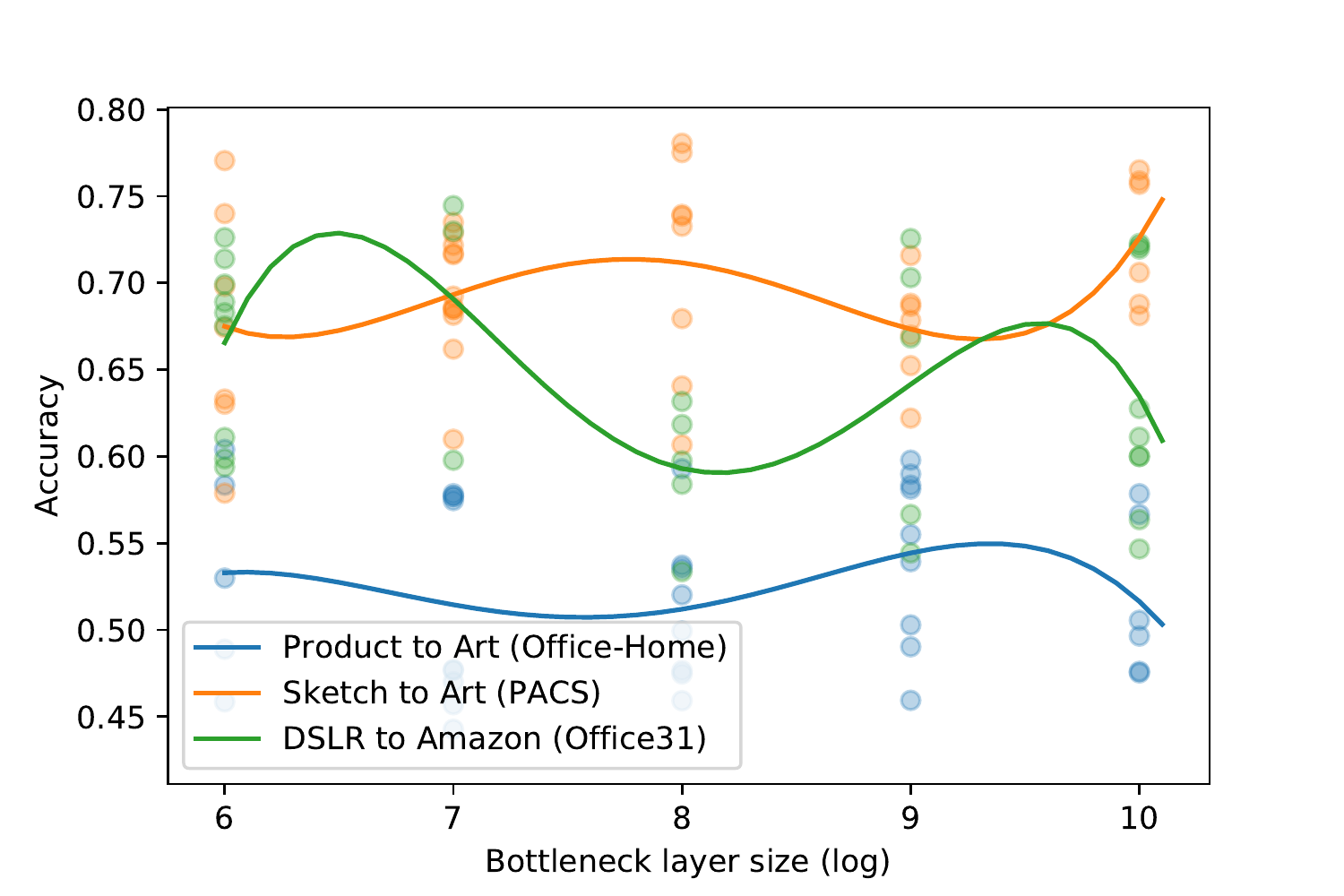}}
    \caption{Search space sensitivity analysis. We randomly sampled $40$ configurations and trained them on a representative setting for each of the considered datasets (Office31, Office-Home, PACS). It is evident that the optimal value changes significantly across tasks. X-axis represents the search space hyper-parameter with respect to which the plot is computed.}
    \label{fig:param_sensitivity}
    \vspace{-1.3em}
\end{figure*}

\subsection{Preliminary analysis}
\label{subsec:prelim_analysis}
\textbf{Search space expressivity and sensitivity.}
Given the added cost of automatically searching for the optimal architecture, it is important to understand whether 1) the design of the adversarial branch has an impact on the final accuracy and 2) if a "good" architecture is actually data-dependent.
To assess this, we followed NAS best practices \cite{Yang2020NASEFH} and randomly sampled $40$ configurations on each setting to evaluate our search space. We observed the following: the design of the adversarial branch can make or break a training. Even excluding a significant ($10\%$) portion of models which failed to converge, on average the design of the adversarial branch accounted for at least $20 \%$ in absolute performance (Fig. \ref{fig:random_sampling} - full results in the supplementary) - note that $\lambda$ was fixed to focus on the architecture only.
Furthermore, as shown in Fig. \ref{fig:param_sensitivity}, the "best" configuration heavily depends on the specific setting being considered. Indeed, while the adversarial branch proposed in ALDA performs well on Office31 and Office-Home (the datasets it was originally tested on), it significantly lags behind on several PACS settings (Fig. \ref{fig:random_sampling} and Table \ref{tab:pacs}).

\textbf{Performance estimation.}
As discussed in sec. \ref{sec:rel_work}, and recently highlighted by \cite{gulrajani2021in}, strong model selection should be a key element of any transfer learning approach; more so when proposing a UDA method capable of explicitly tuning the model hyper-parameters themselves. 
In our next experiments we analyzed how well different metrics correlate with the target accuracy. We sampled $40$ configuration on each setting and computed the metrics discussed in section \ref{subsec:model_selection} at each epoch of training. Full overview on the results can be found in the supplementary, while
 representative plots are shown in Fig. \ref{fig:bimetric_o31_oh}. 
Note how, individually, all metrics perform as weak predictors, correlating with the target accuracy but failing in specific instances. By considering a linear combination we provide a much stronger estimate.

To assess our model selection strategy in isolation from the NAS component, we retrained vanilla ALDA \cite{chen2020adversarial} and reported the performance according to either the last epoch (as done in the paper) or according to our model selection (Tables \ref{tab:office},\ref{tab:office_home},\ref{tab:pacs}). It is worth mentioning that a) we could not replicate their results (even though we used the official code) and b) on Office-Home and PACS our model selection improved average results by up to $2$ full p.p.

\subsection{Comparison with SOTA}
Comparison with current 
sota
methods can be found in  \cref{tab:office_home,tab:pacs,tab:office}. \our significantly improves over the UDA methods it builds on (even when they use our same model selection strategy), with both DANN and ALDA and on all three datasets. The smallest average improvement we observe is $1.9\%$ (Office31 - ALDA $87.5$ vs \our-ALDA $89.4$) while the largest is $3.1\%$ (Office-Home - DANN $56.0$ vs \our-DANN $59.1$). With the exception of Office-Home, where we couldn't replicate the official ALDA results, \our-ALDA is the new sota on all settings.
It is worth mentioning that on one of the easiest Office31 settings  (D-W), \our-ALDA is outperformed by a few different methods; we speculate that, since DSLR and Webcam are very close, the adversarial branch fails to capture meaningful differences. Indeed, ALDA similarly performs poorly on this setting.

\begin{table*}[ht]
    \centering
    \begin{tabular}{lccccccc}
    \hline
                    & A-W & A-D	& D-W &	D-A & W-D &	W-A & AVG \\
                \hline
        ResNet-50 \cite{he2016deep} & 68.4±0.2 & 68.9±0.2 & 96.7±0.1 & 62.5±0.3 & 99.3±0.1 & 60.7±0.3 & 76.1 \\
        DANN \cite{ganin2016domain} & 82.0±0.4 & 79.7±0.4 & 96.9±0.2 & 68.2±0.4 & 99.1±0.1 & 67.4±0.5 & 82.2 \\
        ADDA \cite{tzeng2017adversarial} & 86.2±0.5 & 77.8±0.3 & 96.2±0.3 & 69.5±0.4 & 98.4±0.3 & 68.9±0.5 & 82.9 \\ 
        JAN \cite{long2017deep} & 85.4±0.3 & 84.7±0.3 & 97.4±0.2 & 68.6±0.3 & 99.8±0.2 & 70.0±0.4 & 84.3 \\ 
        MADA \cite{DBLP:conf/aaai/PeiCLW18} & 90.0±0.1 & 87.8±0.2 & 97.4±0.1 & 70.3±0.3 & 99.6±0.1 & 66.4±0.3 & 85.2 \\ 
        CBST \cite{zou2018unsupervised} & 87.8±0.8 & 86.5±1.0 & 98.5±0.1 & 71.2±0.4 & \textbf{100±0.0} & 70.9±0.7 & 85.8 \\ 
        CAN \cite{zhang2018collaborative} & 92.5 & 90.1 & 98.8 & 72.1 & \textbf{100.0} & 69.9 & 87.2 \\ 
        CDAN+E \cite{long2018conditional} & 94.1±0.1 & 92.9±0.2 & 98.6±0.1 & 71.0±0.3 & \textbf{100.0±0.0} & 69.3±0.3 & 87.7 \\ 
        MCS \cite{liang2019distant} & 75.1 & 71.9 & 96.7 & 58.8 & 99.4 & 57.2 & 76.5 \\
        DMRL \cite{wu2020dual} & 90.8±0.3 & 93.4±0.5 & 99.0±0.2 & 73.0±0.3 & \textbf{100.0±0.0} & 71.2±0.3 & 87.9 \\ 
        DM-ADA \cite{xu2020adversarial} & 83.9±0.4 & 77.5±0.2 & \textbf{99.8±0.1} & 64.6±0.4 & 99.9±0.1 & 64.0±0.5 & 81.6 \\
        3CATN \cite{li2019cycle} & 95.3±0.2 & 94.1±0.3 & 99.3±0.5 & 73.1±0.2 & \textbf{100±0.0} & 71.5±0.6 & 88.9 \\
        ALDA \cite{chen2020adversarial} & 95.6±0.5 & 94.0±0.4 & 97.7±0.1 & 72.2±0.4 & \textbf{100.0±0.0} & \textbf{72.5±0.2} & 88.7 \\
        \hline
        ResNet-50  & 67.0±1.9 & 76.4±1.4	& 94.8±1.3	& 56.2±1.0	& 98.7±0.1	& 58.3±0.9	& 75.2  \\
        \vspace{0.2em}
        ResNet-50 + \ourms  & 70.4±2.0 &	77.1±2.0 &	95.6±0.6 &	58.2±2.6 &	99.1±0.5 &	60.1±1.1 &	76.8 \\
        
        DANN  & 85.0±0.5 &	82.5±0.5 &	96.7±0.2 &	63.9±1.1 &	99.2±0.3 &	64.7±0.7 & 82.0  \\
        DANN + \ourms & 85.5±2.1 &	82.8±2.2 &	97.1±1.0 &	64.1±0.5 &	99.8±0.2 &	64.5±2.0 &	82.3  \\
        \vspace{0.2em}
        \our-DANN & 89.4 & 87.6 & 98.4 & 64.1 & 99.8 & 69.3 & 84.8 \\
        
        ALDA  &  94.8±0.6 &	91.6±0.9 &	98.3±0.4 &	69.6±0.9 &	99.9±0.0 &	70.8±0.7 &	87.5 \\ 
        ALDA + \ourms  & 95.0±1.4 &	91.9±0.7 &	98.6±0.4 &	69.3±2.1 &	99.7±0.3 &	71.1±0.8 &	87.6 \\
        \our-ALDA & \textbf{96.1} &	\textbf{95.0} &	98.5 &	\textbf{75.9} &	\textbf{100.0} &	70.7 &	\textbf{89.4}   \\
        \hline
    \end{tabular}
    \caption{Results of \our on Office31 with a ResNet-50 backbone.  + \ourms: experiments run with our model selection.}
    \label{tab:office}
\end{table*}

\begin{table*}
    \centering
    \addtolength{\tabcolsep}{-4pt}
    \begin{tabular}{lccccccccccccc}
    \hline
         & Ar-Cl & Ar-Pr & Ar-Rw & Cl-Ar & Cl-Pr & Cl-Rw & Pr-Ar & Pr-Cl & Pr-Rw & Rw-Ar & Rw-Cl & Rw-Pr & AVG \\ \hline
         ResNet-50 \cite{he2016deep} & 34.9 & 50.0 & 58.0 & 37.4 & 41.9 & 46.2 & 38.5 & 31.2 & 60.4 & 53.9 & 41.2 & 59.9 & 46.1  \\
         DANN \cite{ganin2016domain} & 45.6 & 59.3 & 70.1 & 47.0 & 58.5 & 60.9 & 46.1 & 43.7 & 68.5 & 63.2 & 51.8 & 76.8 & 57.6 \\
        JAN \cite{long2017deep} & 45.9 & 61.2 & 68.9 & 50.4 & 59.7 & 61.0 & 45.8 & 43.4 & 70.3 & 63.9 & 52.4 & 76.8 & 58.3 \\
        CDAN+E \cite{long2018conditional} & 50.7 & 70.6 & 76.0 & 57.6 & 70.0 & 70.0 & 57.4 & 50.9 & 77.3 & \textbf{70.9} & \textbf{56.7} & 81.6 & 65.8 \\
        TAT \cite{liu2019transferable} & 51.6 & 69.5 & 75.4 & 59.4 & 69.5 & 68.6 & 59.5 & 50.5 & 76.8 & \textbf{70.9} & 56.6 & 81.6 & 65.8 \\
        ALDA \cite{chen2020adversarial} & \textbf{53.7} & 70.1 & \textbf{76.4} & \textbf{60.2} & \textbf{72.6} & \textbf{71.5} & 56.8 & \textbf{51.9} & 77.1 & 70.2 & 56.3 & \textbf{82.1} & \textbf{66.6} \\
 \hline
        
        ResNet-50  & 35.6 & 61.5	& 70.4	& 44.0	& 56.1	& 57.4	& 46.5	& 31.9	& 69.1	& 62.4	& 38.0 & 75.6 & 54.1  \\
        \vspace{0.2em}
        ResNet-50 + \ourms  &  36.5 & 63.1 & 71.7 & 45.8 & 	57.5 & 	59.3 & 	47.9 & 	32.3 & 	69.9 & 	63.5 & 	37.3 & 	76.0 & 	55.1 \\ %
        DANN  & 39.8 & 58.0 & 68.1 & 48.6 & 57.0 & 59.9 & 46.9 & 38.4 & 68.8 & 63.2 &	47.7 & 75.9 & 56.0 \\
        DANN + \ourms  &  40.3 & 	60.4 & 	70.1 & 49.6 & 	57.5 & 	60.5 & 	47.9 & 	38.3 & 	69.1 & 	63.7 & 47.7 & 	76.8 & 	56.8 \\
        \vspace{0.2em}
        \our-DANN & 44.9 & 61.1 & 71.2 & 52.7 & 60.4 & 62.5 & 50.1 & 43.1 & 70.0 & 65.4 & 50.9 & 77.1 & 59.1 \\ %
        ALDA  & 46.4 & 68.6 & 74.6 & 57.6 & 67.0 & 69.4 & 57.2 & 46.3 & 75.6 & 69.2 & 53.3 &	80.8 & 63.8 \\
        ALDA + \ourms &  47.5 & 	70.1 & 	75.2 & 58.8 & 	67.5 & 	69.7 & 	58.1 & 	46.6 & 	76.2 & 	69.9 & 	54.0 & 	81.0 & 	64.6 \\
        \our-ALDA &  51.5 & \textbf{71.7} & 75.5 & 	59.8 & 	69.4 & 	69.5 & 	\textbf{59.8} & 	47.1 & 	\textbf{77.7} & 	70.6 & 	55.2 & 	80.2 & 	65.7  \\
        \hline
    \end{tabular}
    \caption{Results of \our on Office-Home, using a ResNet-50 backbone. Note we were not able to reproduce the official ALDA numbers using the official code, + \ourms: experiments run with our model selection strategy. }
    \label{tab:office_home}
\end{table*}	

\begin{table*}[t]
    \centering
    \addtolength{\tabcolsep}{-2pt}
    \begin{tabular}{lccccccccccccc}
    \hline
         & P-A & C-A & S-A & A-P & C-P & S-P & A-C & S-C & P-C & A-S & C-S & P-S & AVG \\ \hline
        ALDA \cite{chen2020adversarial} & 89.3	& 91.9 & 69.9	& 98.3 & 97.3 &	63.4 &	85.1 &	75.2 &	\textbf{74.3} &	79.2 &	70.6 &	60.7 &	79.6  \\ 
        ALDA + \ourms  &  90.2 &	\textbf{92.0} &	72.3 &	98.4 &	\textbf{97.8} & 69.5 &	86.0 &	82.1 &	72.1 &	\textbf{80.7} &	\textbf{75.1} &	\textbf{66.1} &	81.9 \\ 
        \our-ALDA  & \textbf{93.1} & 	91.8 & 	\textbf{78.1} & 	\textbf{98.7} & 	\textbf{97.8} & 	\textbf{70.8} & 	\textbf{88.7} & 	\textbf{84.9} & 	69.7 & 	79.8 & 	69.5 & 	64.9 & 	\textbf{82.3}  \\ \hline
    \end{tabular}
    \caption{Results of \our on PACS, using a ResNet-50 backbone, + \ourms: experiments run with our model selection strategy.  Due to space constraints, ResNet50 and DANN experiments are in the supplementary materials.}
    \label{tab:pacs}
\end{table*}

\section{Discussion}
\label{sec:discussion}
The shift from hand-crafted 
to %
principled architecture search is ongoing, as it is witnessed by the flourishing NAS research and techniques.
\our fills in two important requirements of NAS for UDA: it provides a data-driven strategy (\ourms) for model selection, circumventing the lack of target labels; and it focuses on searching the architecture of auxiliary branches attached to a pre-trained backbone, essential practise for state-of-the-art performance.

Only most recent research has stressed %
the importance of model selection for UDA~\cite{gulrajani2021in}.
The lack of a validation identically distributed as the test set,  led researchers to indulge on selecting hyper-parameters on the test set.
Hopefully the propositions here (cf.\ \ourms) would be initial steps towards more methodologically sound model selection practises. 

Finally, we emphasize that \our provides tools (search space, \ourms, branch optimization), which comply with the use of most other NAS methods and potentially all UDA techniques based on an auxiliary branch.
A notable example is self-supervision-based UDA methods using an auxiliary branch. %
Since they adopt the same exact macro-architecture as in Fig. \ref{fig:da_branches}, then \our seamlessly applies there.

\section*{Acknowledgments}
We acknowledge the CINECA award under the \mbox{ISCRA} initiative, for the availability of high performance computing resources and support. Special thanks go to Giuseppe Fiameni (NVIDIA AI Technology Centre, Italy) for his insights and generous technical support. The work was partially supported by the ERC project N. 637076 RoboExNovo and the research herein was carried out using the IIT HPC infrastructure. This work was supported by the CINI Consortium through the VIDESEC project.

\clearpage
\clearpage

{\small
\bibliographystyle{ieee_fullname}
\bibliography{egbib}
}

\clearpage
\title{Adversarial Branch Architecture Search for Unsupervised Domain Adaptation\\Supplementary Materials}
\predate{}
\postdate{}
\date{}
\maketitle
\thispagestyle{empty}
\setcounter{section}{0}
\setcounter{equation}{0}
\setcounter{table}{0}
\setcounter{figure}{0}

\newcommand{\theHalgorithm}{\arabic{algorithm}}
We provide additional analysis, results and implementation details, to further support the claims of the paper and illustrate specific insights. In detail, we present here:
\begin{itemize}
    \item 
    A more in-depth description of how the Ensemble Model Selection (EMS) module is trained, as well as further implementation details, plots and experimental results regarding it and the different metric it builds on for target accuracy prediction [sec. \ref{app:estimators}].
    \item 
    An evaluation of the impact of the search space parameters on the final performance of the UDA method [sec. \ref{app:search}].
    \item 
    A description of our ResNet50 and ResNet50 + EMS baselines. [sec. \ref{app:baseline}]
    \item 
    The full PACS results [sec. \ref{app:pacs}].
\end{itemize}

\section{Ensemble Model Selection (EMS)}
\label{app:estimators}
EMS is a crucial component of the \our pipeline: in the UDA setting the target labels are not available, and NAS cannot be applied to UDA without a way to assess the quality of the proposed solutions. To be more explicit, EMS uses the metrics discussed in section 3.3 to provide feedback on how good a sample is.

\subsection{Training EMS}
In this work we build an ensemble of weak predictors by training a linear regressor on top of label-free metrics which weakly correlate with the target accuracy.
To do so, we randomly sampled $200$ configurations on each dataset and trained them while collecting $100$ snapshots for each run, thus ending with a dataset of $20,000$ points. At each snapshot we collected the $6$ metrics described in sec. 3.3 thus building a dataset of size $2000\times6$ on which to train our regressor; note that the pseudo-label metric is only used to assess the best snapshot in a single run, not to compare across runs. All input features were standardized to zero mean and unit standard deviation. To model the regressor we used a linear least-squares model as implemented in Scikit-Learn \cite{scikit-learn}. As target labels (and thus the corresponding accuracy) is not available in the context of UDA, we trained the regressor on a different dataset and transferred it to the one of interest. Specifically, on Office31 we used the regressor trained on Office-Home and vice-versa. The regressor for PACS was trained on Office31.

\subsection{Unsupervised performance estimators} 
\begin{figure*}
    \centering
    \includegraphics[width=0.99\textwidth]{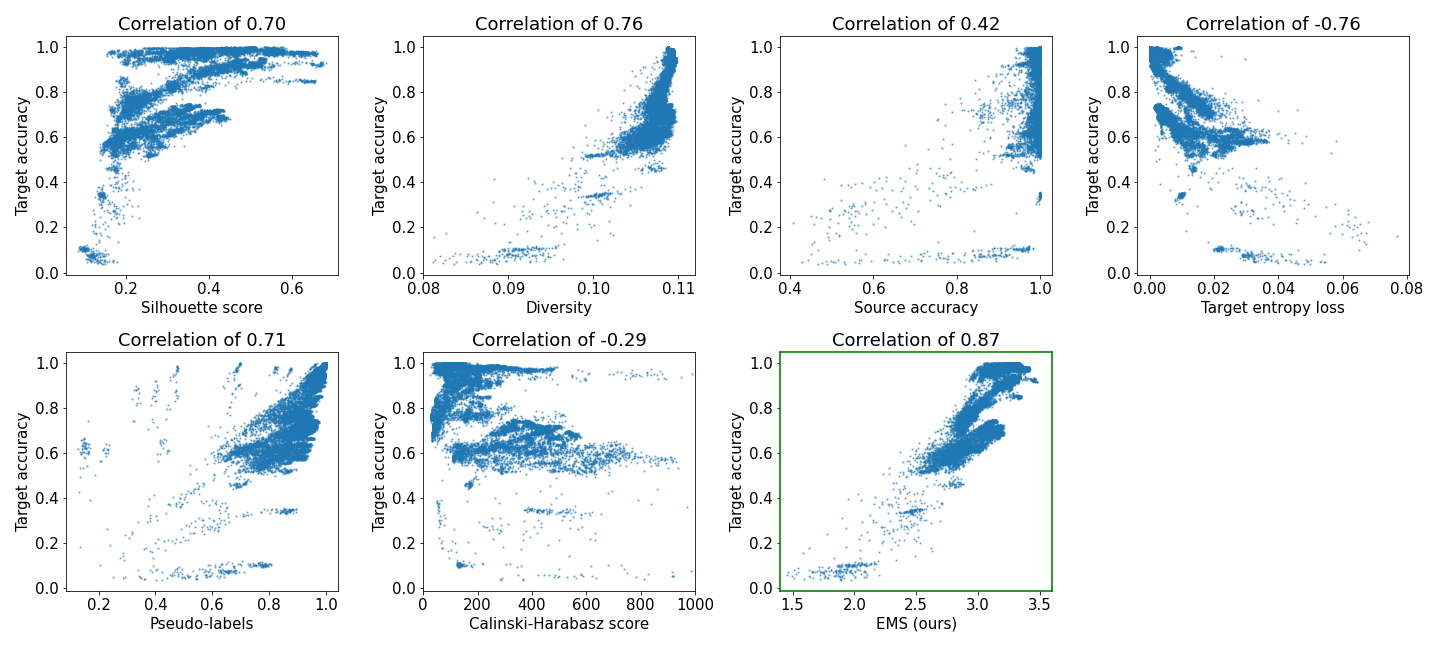}
    \caption{Correlation between the metrics introduced in section 3.3 and the target accuracy, as computed on Office31.}
    \label{fig:metric_correlation_o31}
\end{figure*}

\begin{figure*}
    \centering
    \includegraphics[width=0.99\textwidth]{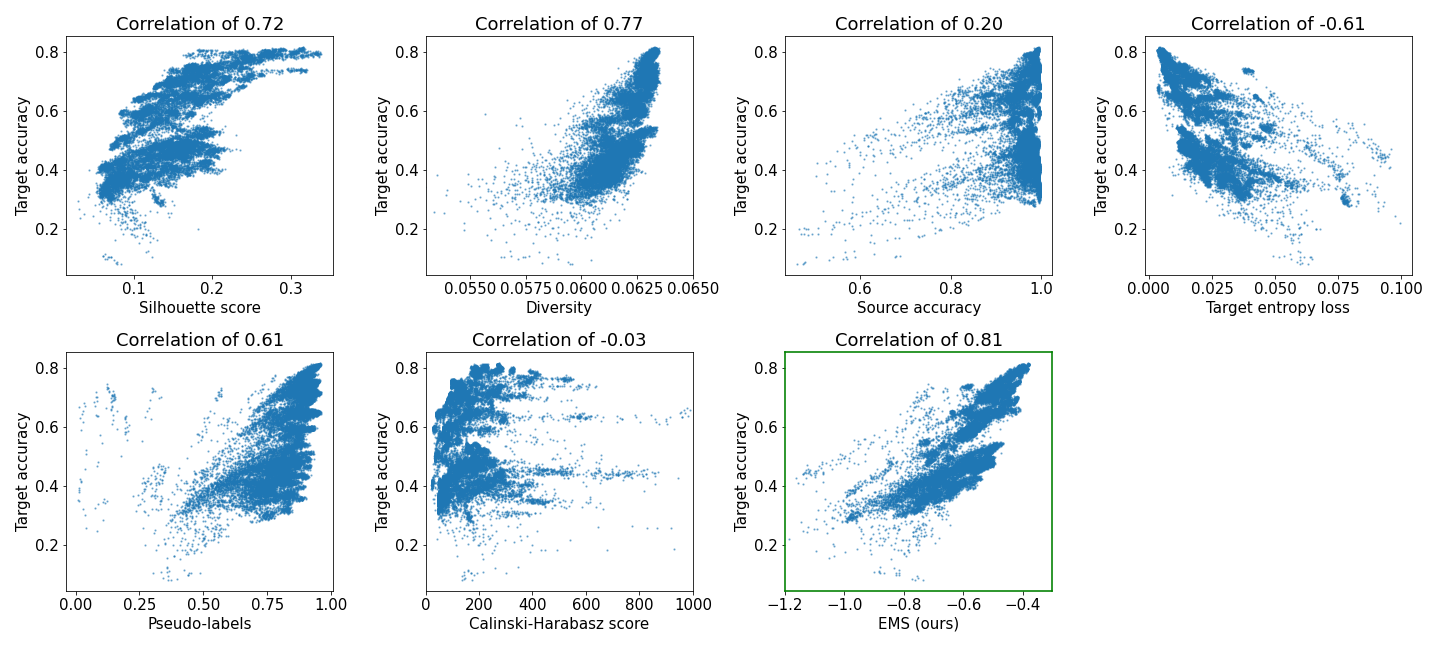}
    \caption{Correlation between the metrics introduced in section 3.3 and the target accuracy, as computed on Office-Home.}
    \label{fig:metric_correlation_oh}
\end{figure*}

Figures \ref{fig:metric_correlation_o31}, \ref{fig:metric_correlation_oh} and Table \ref{tab:metric_correlation} (in aggregated form) show how well the predictors, presented in section 3.3 of the main paper, perform. EMS achieves consistently better correlation on average and future work could investigate if more complex model ensembles could achieve even better performance. Note that since the regressors were only trained on Office31 and Office-Home, Table \ref{tab:search_space} and the corresponding figures are focused on those two datasets.

\subsection{Implementation of the Diversity metric}
The original Diversity \cite{wu2020entropy} is defined as:
\begin{equation}
    H(\hat{q}(\mathcal{T})) = - \sum_{k=1}^{K} \hat{q}_k log(\hat{q}_k)
\end{equation}
where $K$ is the number of classes. As we use the Diversity as a predictor for estimating performance across settings with different values of $K$, we simply divide the metric by $K$ as to get an average over the classes:
\begin{equation}
    H(\hat{q}(\mathcal{T})) = - \frac{1}{K}\sum_{k=1}^{K} \hat{q}_k log(\hat{q}_k)
\end{equation}

\begin{table}[htbp]
    \centering
    \begin{tabular}{lccc}
    \toprule
              & \multicolumn{3}{c}{Average corr. with Target Acc.} \\
        Metric &  Office31 & Office-Home & PACS\\
        \midrule
        Entropy & 0.76 & 0.61 & 0.57\\
        Diversity & 0.76 & 0.77 & 0.76\\
        Pseudo-Labels & 0.71 & 0.61 & 0.68\\
        Source Accuracy & 0.42 & 0.20 & 0.48\\
        Silhouette & 0.70 & 0.72 & 0.74\\
        Calinski-Harabasz & 0.29 & 0.03 & -0.13\\
        \midrule
        \ourms & \textbf{0.87} & \textbf{0.81} & \textbf{0.80}\\
        \bottomrule
    \end{tabular}
    \caption{This table shows how the different metrics introduced in section 3.3 of the main paper correlate with the target accuracy. Note that all metrics, with the exception of source accuracy are computed on the target.} %
    \label{tab:metric_correlation}
\end{table}

\section{Impact of the auxiliary branch configuration on the final accuracy}
\label{app:search}
As can be seen in Table \ref{tab:search_space}, the design choices which are part of our search space have a tremendous impact on the final performance of an ALDA \cite{chen2020adversarial} training - in other words, the architecture and the hyper-parameters of the auxiliary branch can make or break the UDA approach that builds on it.

\begin{table*}[!htbp]
    \centering
    \begin{tabular}{lccccc}
    \toprule
         \multirow{2}{*}{Setting} & \% & \multirow{2}{*}{Min} & \multirow{2}{*}{Max} & Mean  & \multirow{2}{*}{Range} \\
         & Div. & & & $\pm$ Std. & \\
         \midrule
        A-W & 0.23 & 64.59 & 95.79 & $83.4 \pm 09.8$ & 31.19 \\
        A-D & 0.28 & 72.49 & 91.57 & $84.4 \pm 06.2$ & 19.08 \\
        D-W & 0.15 & 17.11 & 99.06 & $89.0 \pm 21.7$ & 81.95 \\
        D-A & 0.28 & 46.59 & 70.15 & $59.3 \pm 06.4$ & 23.55 \\
        W-D & 0.22 & 16.47 & 100.00 & $89.3 \pm 24.2$ & 83.53 \\
        W-A & 0.17 & 34.97 & 74.55 & $63.8 \pm 08.3$ & 39.58 \\
        Ar-Cl & 0.08 & 26.60 & 50.71 & $41.7 \pm 06.4$ & 24.11 \\
        Ar-Pr & 0.14 & 59.12 & 65.14 & $63.2 \pm 02.1$ & 06.01 \\
        Ar-Rw & 0.22 & 43.30 & 75.89 & $70.0 \pm 06.3$ & 32.59 \\
        Cl-Ar & 0.14 & 36.55 & 57.87 & $51.6 \pm 07.3$ & 21.32 \\
        Cl-Pr & 0.14 & 55.58 & 66.41 & $60.9 \pm 03.9$ & 10.84 \\
        Cl-Rw & 0.15 & 46.12 & 70.25 & $63.2 \pm 05.6$ & 24.13 \\
        Pr-Ar & 0.14 & 40.15 & 57.21 & $49.8 \pm 04.9$ & 17.06 \\
        Pr-Cl & 0.11 & 26.27 & 49.52 & $39.7 \pm 06.5$ & 23.25 \\
        Pr-Rw & 0.07 & 41.47 & 75.64 & $66.5 \pm 10.7$ & 34.16 \\
        Rw-Ar & 0.07 & 59.81 & 67.86 & $64.7 \pm 02.3$ & 08.06 \\
        Rw-Cl & 0.07 & 15.40 & 54.80 & $46.2 \pm 07.6$ & 39.40 \\
        Rw-Pr & 0.14 & 70.30 & 81.36 & $77.1 \pm 02.6$ & 11.06 \\
        \midrule
        AVG & 0.2 & 42.9 & 72.4 & $64.6 \pm 7.9$ & 29.49 \\
        \bottomrule 
    \end{tabular}
    \caption{Over $400$ different points were randomly sampled from our search space and the corresponding models were then trained using ALDA \cite{chen2020adversarial} over either Office31 \cite{saenko2010adapting} (top 6 rows) or Office-Home \cite{venkateswara2017deep} settings. Note how the configuration of the auxiliary branch significantly affects the model performance. \textit{\% Div.} stands for the fraction of runs which did not converge - \textit{Min} is the minimum accuracy - \textit{Max} is the maximum -  \textit{Mean} reports the mean $\pm$ the standard deviation, and \textit{Range} is the total range of accuracies, defined as $Max - Min$.%
    }
    \label{tab:search_space}
\end{table*}

\section{Baseline implementation}
\label{app:baseline}
For all our baselines, we use a ResNet-50 backbone pre-trained on ImageNet.
The DANN and ALDA baselines use the same adversarial branch architecture (two 1024 fully connected layers), connected right after a 512-sized bottleneck layer preceding the ResNet-50 output.
The initial learning rate is set to 0.001, and it is adjusted during training following
\begin{equation}
     \mu_p = \frac{\mu_0}{(1 + \alpha \cdot p)^{\beta}},
\end{equation}
where $\alpha=10$, $\beta=0.75$ and $p$ grows from 0 to 1 during training.
The weight decay is 0.0005, and the network is trained with SGD (momentum 0.9, batch size 36).
Since the bottleneck and adversarial branch are trained from scratch, for those parts of the network we set the initial learning rate to 0.01 and the weight decay to 0.001.
For each run (no EMS) we select the last two snapshots (evaluated each 100 iterations) and average them.
For the EMS baselines we select the best snapshot of the run according to the value predicted by the EMS regressor.
Each run was repeated 4 times and the final accuracies were averaged.
\section{PACS results}
\label{app:pacs}

Table \ref{tab:pacs_suppl} did not fully fit in the main text and is here reported in its entirety. ABAS manages to significantly improve over both DANN and ALDA: surprisingly, on Sketch to Photo, ABAS-DANN largely outperforms ABAS-ALDA, highlighting how an older method can perform better than a new one, when trained with the optimal branch.
Indeed, although in a specific setting (P-S), our EMS drives BOHB to select a less than optimal architecture for DANN, ABAS-DANN offers a slightly superior performance overall. Future work to improve EMS would be expected to further increase our results with both methods.

\begin{table*}[!htbp]
    \centering
    \addtolength{\tabcolsep}{-1pt}
    \begin{tabular}{lccccccccccccc}
    \hline
         & P-A & C-A & S-A & A-P & C-P & S-P & A-C & S-C & P-C & A-S & C-S & P-S & AVG \\ \hline
        ResNet-50 \cite{he2016deep} &  62.3 & 71.2 & 30.8 & 96.6 & 89.7 & 38.0 & 57.0 & 53.5 & 27.6 & 43.0 & 59.1 & 30.7 & 55.0  \\
        \vspace{0.2em}
        ResNet-50 + \ourms  &  62.0	& 72.0	& 25.6	& 97.7	& 90.9 & 38.8 & 57.5 & 53.2 & 23.0 & 41.3 & 58.7 & 28.2 & 54.1 \\ 
        DANN \cite{ganin2016domain} &  79.9	& 88.4	& 70.3	& 97.3	& 95.1	& 57.1	& 82.4	& 70.8	& 64.7	& 62.5	& 68.7	& 61.4	& 74.9  \\
        DANN + \ourms  &  80.3	& 89.0	& 72.6	& 98.1	& 95.6	& 56.2	& 82.2	& 73.3	& 64.0	& 63.4	& 69.7	& 59.6	& 75.3 \\
         \vspace{0.2em}
        \our-DANN & 91.7	& 90.7	& 68.2	& 98.3	& 96.2	& \textbf{82.8}	& 87.2	& 71.2	& 73.7	& 68.7	& 76.0	& 47.2	& \textbf{82.9}\\ 
        ALDA \cite{chen2020adversarial} & 89.3	& 91.9 & 69.9	& 98.3 & 97.3 &	63.4 &	85.1 &	75.2 &	\textbf{74.3} &	79.2 &	70.6 &	60.7 &	79.6  \\ 
        ALDA + \ourms  &  90.2 &	\textbf{92.0} &	72.3 &	98.4 &	\textbf{97.8} & 69.5 &	86.0 &	82.1 &	72.1 &	\textbf{80.7} &	\textbf{75.1} &	\textbf{66.1} &	81.9 \\ 
        \our-ALDA  & \textbf{93.1} & 	91.8 & 	\textbf{78.1} & 	\textbf{98.7} & 	\textbf{97.8} & 	70.8 & 	\textbf{88.7} & 	\textbf{84.9} & 	69.7 & 	79.8 & 	69.5 & 	64.9 & 	82.3  \\ \hline
    \end{tabular}
    \caption{Results of \our on PACS, using a ResNet-50 backbone, + \ourms: experiments run with our model selection strategy.}
    \label{tab:pacs_suppl}
\end{table*}

\section{Software and hardware}
Our network training code is written in PyTorch and based on the publicly available ALDA \cite{chen2020adversarial} repository, likewise, we use HpBandSter, the official  
BOHB \cite{falkner2018bohb} implementation.
The BOHB algorithm is run for $24$ iterations using $8$ parallel workers, each running on a Tesla V100 GPU. 
On average, it takes $6$, $10$ and $6.4$ hours to run the full \our pipeline on PACS, Office-Home and Office31 respectively, leading to a cost of $80$ GPU hours on the largest setting.

\clearpage
\clearpage
{\small
\bibliographystyle{ieee_fullname}
\bibliography{egbib}
}

\end{document}